%% file: arxiv.tex
\documentclass[final]{cvpr}

\usepackage{times}
\usepackage{epsfig}
\usepackage{graphicx}
\usepackage{amsmath}
\usepackage{amssymb}
\usepackage{pifont}
\newcommand{\xmark}{\ding{55}}%
\newcommand{\cmark}{\ding{51}}%

\usepackage{makecell}
\usepackage{booktabs} 
\usepackage{multirow}
\usepackage{color, xcolor}
\usepackage{tabularx,verbatim}
\usepackage{xspace}
\usepackage{caption}

\usepackage[ruled,vlined]{algorithm2e}
\SetArgSty{textnormal}

\let\oldnl\nl
\newcommand{\nonl}{\renewcommand{\nl}{\let\nl\oldnl}}

\usepackage{setspace}
\usepackage{comment}
\usepackage{color, colortbl}

\definecolor{Gray}{gray}{0.92}
\newcolumntype{L}{>{\arraybackslash}m{2cm}}
\DeclareMathOperator*{\Max}{max}
\DeclareMathOperator*{\ArgMax}{argmax}
\DeclareMathOperator*{\Argmin}{argmin}
\DeclareMathOperator*{\Min}{min}
\DeclareMathOperator{\E}{\mathbb{E}}
\DeclareMathOperator*{\under}{\mathbb{E}}
\newcommand{\ours}{CGCT\xspace}
\newcommand{\domours}{D-CGCT\xspace}
\newcommand{\dcl}{DCL\xspace}

\usepackage[labelsep=period]{caption}

\usepackage[pagebackref=true,breaklinks=true,colorlinks,bookmarks=false]{hyperref}

\colorlet{dark-blue}{blue!70!black}
\hypersetup{
    colorlinks=true,%
    citecolor=dark-blue,%
    filecolor=dark-blue,%
    linkcolor=red,%
    urlcolor=magenta
}

\def\confYear{CVPR 2021}
\definecolor{mypink}{RGB}{219, 48, 122}

\begin{document}

\title{Curriculum Graph Co-Teaching for Multi-Target Domain Adaptation}

\author{Subhankar Roy$^{\textcolor{mypink}{1,2}}\thanks{Equal contribution}$~, Evgeny Krivosheev$^{\textcolor{mypink}{1}*}$, Zhun Zhong$^{\textcolor{mypink}{1}}\thanks{Corresponding author}$~, Nicu Sebe$^{\textcolor{mypink}{1}}$, Elisa Ricci$^{\textcolor{mypink}{1,2}}$ \\
 \large{$^{\textcolor{mypink}{1}}$University of Trento,  Italy~~$^{\textcolor{mypink}{2}}$Fondazione Bruno Kessler, Italy}\\
 \small{Project Page: \url{https://roysubhankar.github.io/graph-coteaching-adaptation}}
}

\maketitle

\begin{abstract}
In this paper we address multi-target domain adaptation (MTDA), where given one labeled source dataset and multiple unlabeled target datasets that differ in data distributions, the task is to learn a robust predictor for all the target domains. We identify two key aspects that can help to alleviate multiple domain-shifts in the MTDA: feature aggregation and curriculum learning. To this end, we propose Curriculum Graph Co-Teaching (CGCT) that uses a dual classifier head, with one of them being a graph convolutional network (GCN) which aggregates features from similar samples across the domains. To prevent the classifiers from over-fitting on its own noisy pseudo-labels we develop a co-teaching strategy with the dual classifier head that is assisted by curriculum learning to obtain more reliable pseudo-labels. Furthermore, when the domain labels are available, we propose Domain-aware Curriculum Learning (DCL), a sequential adaptation strategy that first adapts on the easier target domains, followed by the harder ones. We experimentally demonstrate the effectiveness of our proposed frameworks on several benchmarks and advance the state-of-the-art in the MTDA by large margins (e.g. +5.6\% on the DomainNet).
\end{abstract}
\vspace{-4mm}
\input{intro.tex}

\input{related.tex}



\input{method}

\input{experiments.tex}

\section{Conclusion}
To address multi-target domain adaptation (MTDA), we proposed Curriculum Graph Co-Teaching (CGCT) that uses a graph convolutional network to perform robust feature aggregation across multiple domains, which is then trained with a co-teaching and curriculum learning strategy. To better exploit domain labels of the target we presented a Domain-aware curriculum (DCL) learning strategy that adapts easier target domains first and harder later, enabling a smoother feature alignment. Through extensive experiments we demonstrate that our proposed contributions handsomely outperform the state-of-the-art in the MTDA.\\
{\noindent\textbf{Acknowledgements}} This work is supported by the EU H2020 SPRING No. 871245 and AI4Media No. 951911 projects; the Italy-China collaboration project TALENT:2018YFE0118400; and the Caritro Deep Learning Lab of the ProM Facility of Rovereto.

\newpage

\renewcommand{\thesection}{\Alph{section}}
\setcounter{section}{0}

\begin{center}
    \Large\textbf{Supplementary Material}
\end{center}

\input{supp}

{\small
\bibliographystyle{ieee_fullname}
\bibliography{egbib}
}

\end{document}

%% file: intro.tex
\section{Introduction}
\label{sec:intro}


Deep learning models suffer from the well known drawback of failing to generalize well when deployed in the real world. The gap in performance arises due to the difference in the distributions of the training (a.k.a source) and the test (a.k.a target) data, which is popularly referred to as \textit{domain-shift}~\cite{torralba2011unbiased}. Since, collecting labeled data for every new operating environment is prohibitive, a rich line of research, called Unsupervised Domain Adaptation (UDA), has evolved to tackle the task of leveraging the source data to learn a robust predictor on a desired target domain. 

In the literature, UDA methods have predominantly been designed to adapt from a single source domain to a single target domain (STDA). Such methods include optimizing statistical moments~\cite{tzeng2014deep, long2015learning, sun2016deep, peng2018synthetic, carlucci2017autodial, carlucci2017just, chang2019domain, roy2019unsupervised}, adversarial training~\cite{ganin2016domain, Hoffman:Adda:CVPR17, long2018conditional}, generative modelling~\cite{russo17sbadagan, hoffman2017cycada, liu2016coupled}, to name a few. However, given the proliferation in unlabeled data acquisition, the need to adapt to just a single target domain has lost traction in the real world scenarios. As the number of target domains grows, the number of models that need to be trained also scales linearly. For this reason, the research focus has very recently been steered to address a more practical scenario of adapting simultaneously to multiple target domains from a single source domain. This adaptation setting is formally termed as Multi-target Domain Adaptation (MTDA). The goal of the MTDA is to learn more compact representations with a single predictor that can perform well in all the target domains. Straightforward application of the STDA methods for the MTDA may be sub-optimal due to the presence of multiple domain-shifts, thereby leading to negative transfer~\cite{zhang2020overcoming, chen2019blending}. Thus, the desideratum to align multiple data distributions makes the MTDA considerably more challenging. 

In this paper we build our framework for the MTDA pivoted around two key concepts: \textit{feature aggregation} and \textit{curriculum learning}. Firstly, we argue that given the intrinsic nature of the task, learning robust features in a unified space is a prerequisite for attaining minimum risk across multiple target domains. For this purpose we propose to represent the source and the target samples as a graph and then leverage Graph Convolutional Networks~\cite{kipf2017semi} (GCN) to aggregate semantic information from similar samples in a \textit{neighbourhood} across different domains. For the GCN to be operative, partial relationships among the samples (nodes) in the graph must at least be known apriori in the form of class labels. However, this information is absent for the target samples. To this end, we design a \textit{co-teaching} framework where we train two classifiers: a MLP classifier and a GCN classifier that provide target pseudo-labels to each other. On the one hand, the MLP classifier is utilized to make the GCN learn the pairwise similarity between two nodes in the graph. While, on the other hand, the GCN classifier, due to its feature aggregation property, provides better pseudo-labels to assist the training of the MLP classifier. Given that co-teaching works on the assumption that different networks capture different aspects of learning~\cite{blum1998combining}, it is beneficial for suppressing noisy pseudo-labels. 
his feature aggregation and/or co-teaching aspects are largely missing in existing MTDA methods~\cite{chen2019blending, gholami2020unsupervised, peng2019domain, yang2020heterogeneous} (see Tab.~\ref{tab:mtda_table}).

\begin{table}[!t]
    \centering
    \footnotesize
    \setlength{\tabcolsep}{4.5pt}
    \begin{tabular}{lcccc}
         \hline
         \thead{Method}& \thead{\makecell{Domain \\labels}} & \thead{\makecell{Feature \\aggregation}} & \thead{\makecell{Curriculum \\learning}} & \makecell{Co-\\teaching} \\
         \hline
         AMEAN~\cite{chen2019blending} & \xmark & \xmark & \xmark & \xmark \\
         DADA~\cite{peng2019domain} & \xmark & \xmark & \xmark & \xmark\\
         MTDA-ITA~\cite{gholami2020unsupervised} & \cmark & \xmark & \xmark & \xmark \\
         HGAN~\cite{yang2020heterogeneous} & \cmark & \cmark & \xmark & \xmark\\
         \hline
         CGCT (\textbf{Ours}) & \xmark & \cmark & \cmark & \cmark \\
         D-CGCT (\textbf{Ours}) & \cmark & \cmark & \cmark & \cmark \\
         \hline
    \end{tabular}
    \vspace{-.12in}
    \caption{Comparison with recent the state-of-the-art MTDA methods in terms of the operating regimes.}
    \label{tab:mtda_table}
    \vspace{-4.5mm}
\end{table}

Secondly, we make a crucial observation, very peculiar to the MTDA setting, \textit{i.e.}, during training as the network tries to adapt to multiple domain-shifts of varying degree, pseudo-labels obtained on-the-fly from the network for the target samples are very noisy. Self-training the network with unreliable pseudo-labeled target data further deteriorates the performance. To further combat the impact of noisy pseudo-labels, we propose to obtain pseudo-labels in an episodic fashion, and advocate the use of \textit{curriculum learning} in the context of MTDA. In particular, when the domain labels of the target are latent, each episode or \textit{curriculum step} consists of a fixed number of training iterations. Fairly consistent and reliable pseudo-labels are obtained from the GCN classifier at the end of each curriculum step. We call this proposed framework as \textbf{C}urriculum \textbf{G}raph \textbf{C}o-\textbf{T}eaching (\ours) (see Fig.~\ref{fig:flow} (a)).

Furthermore, when the domain labels of the target are available, we propose an Easy-To-Hard Domain Selection (EHDS) strategy where the feature alignment process begins with the target domain that is closest to the source and then gradually progresses towards the hardest one. This makes adaptation to multiple targets smoother. In this case, each curriculum step involves adaptation with a single new target domain. The \ours when combined with this proposed \textbf{D}omain-aware \textbf{C}urriculum \textbf{L}earning (DCL)  (see Fig.~\ref{fig:flow} (b)) is referred to as \domours. The Tab.~\ref{tab:mtda_table} highlights the operating regimes of our frameworks versus the state-of-the-art MTDA methods. To summarize, the contributions of this work are threefold:
\begin{itemize}
    \vspace{-2mm}
    \item We propose Curriculum Graph Co-Teaching (\ours) for MTDA that exploits the co-teaching strategy with the dual classifier head, together with the curriculum learning, to learn more robust representations across multiple target domains.
    \vspace{-2mm}
    \item To better utilize the domain labels, we propose a Domain-aware Curriculum Learning (DCL) strategy to make the feature alignment process smoother.
    \vspace{-2mm}
    \item In the MTDA setting, we outperform the state-of-the-art for several UDA benchmarks by significant margins (including +5.6\% on the large scale DomainNet~\cite{peng2019moment}).
\end{itemize}

%% file: related.tex
\vspace{-3mm}
\section{Related Works}
\label{sec:related}
\vspace{-2mm}
\textbf{Single-source single-target DA} (STDA) refers to the task of adapting a classifier from a single labeled source dataset to a single unlabeled target dataset. In the UDA literature, a plethora of STDA methods have been proposed, which can be broadly classified into three major categories based upon the adaptation strategy. The first category uses first (Maximum Mean Discrepancy ~\cite{tzeng2014deep, long2015learning, long2016deep, venkateswara2017Deep}) or second order (correlation alignment~\cite{sun2016deep, morerio2017minimal, peng2018synthetic, li2016revisiting, carlucci2017autodial, chang2019domain, mancini2018boosting, roy2019unsupervised, roy2019_full_wc_unsupervised}) statistics of the source and target features to align the marginal feature distributions.
%
%
%
%
The second category of STDA methods~\cite{ganin2016domain, Hoffman:Adda:CVPR17, cao2018partial, long2018conditional, chen2018re} adopts adversarial training strategy to align the marginal feature distributions of the two domains. Essentially, these methods use a gradient reversal layer~\cite{ganin2016domain} to make the feature extractor network agnostic to domain specific information. The final category of STDA methods~\cite{russo17sbadagan, hoffman2017cycada, sankaranarayanan2018generate, liu2016coupled} resort to pixel-level adaptation by generating synthetic \textit{target-like} source images or \textit{source-like} target images with the help of generative adversarial network (GAN)~\cite{goodfellow2014generative}. However, practical applications go beyond the single-source and single-target setting and often involve multiple source~\cite{xu2018deep, roy2020trigan, yang2020curriculum} or target domains.

\textbf{Multi-target DA} aims to transfer knowledge from a single labeled source dataset to multiple unlabeled target datasets. 
While the research in STDA is quite mature, most STDA methods can not be trivially extended to a multi-target setting. 
So far only a handful of methods~\cite{chen2019blending, peng2019domain, jin2020minimum, liu2020open, gholami2020unsupervised, yang2020heterogeneous} for MTDA can be found in the literature. 
AMEAN~\cite{chen2019blending} performs clustering on the blended target domain samples to obtain \textit{sub-targets} and then learns domain-invariant features from the source and the obtained sub-targets using a STDA method~\cite{shu2018dirt}. The approaches introduced in~\cite{peng2019domain, jin2020minimum, gholami2020unsupervised} are derived from STDA and do not exploit any peculiarity of the MTDA setting. Conversely, our \ours and \domours are tailor-made for the multi-target setting as we propose to use feature aggregation of similar samples across multiple domains.

\textbf{Curriculum for DA} involves adopting an adaptive strategy that evolves over time to better address the adaptation across domains. Shu \textit{et. al.}~\cite{shu2019transferable} propose a strategy based on curriculum learning that exploits the loss of the network as weights to identify and eliminate unreliable source samples. An Easy-to-Hard Transfer Strategy (EHTS) is proposed in PFAN~\cite{chen2019progressive} that progressively selects the pseudo-labeled target samples which have higher cosine similarity to the per-category source prototypes. Similarly, our \ours is inspired by the EHTS strategy except we progressively recruit the pseudo-labeled targets~\cite{bengio2009curriculum} from the robust GCN classification head to better train the MLP classifier, which in turn regularizes the GCN head (see Sec.\ref{sec:cgct}). For the multi-source DA setting, CMSS~\cite{yang2020curriculum} trains a separate network to weigh the most relevant samples across several source domains for adapting to a single target domain. However, differently from CMSS, our proposed DCL utilizes the domain information to adapt over time from the easiest to the hardest target domain in the MTDA setting (see Sec.~\ref{sec:domain_curr}).

\textbf{Graph Neural Networks} (GNN) are neural network models applied on graph-structured data that can capture the relationships between the objects (nodes) in a graph via message passing through the edges~\cite{gori2005new, wu2020comprehensive}. Relevant to our work are 
GNN-derived Graph Convolutional Networks (GCN)~\cite{kipf2017semi} that have recently been applied for addressing DA~\cite{ma2019gcan, luo2020progressive, yang2020heterogeneous}. For instance, Luo \textit{et. al.}~\cite{luo2020progressive} propose PGL for open-set DA to capture the relationship between the overlapping classes in the source and the target.
Notably, Yang \textit{et. al.}~\cite{yang2020heterogeneous} introduce heterogeneous Graph Attention Network (HGAN) for MTDA to learn the relationship of similar samples among multiple domains and then utilize the graph-based pseudo-labeled target samples to align their centroids with that of the source. Unlike ~\cite{luo2020progressive, yang2020heterogeneous}, we incorporate the idea of co-teaching~\cite{han2018co} in a GCN framework for combating noisy pseudo-labels.

%% file: method.tex
\vspace{-1mm}
\section{Method}
\label{method}
\vspace{-2mm}
In this section we present our proposed Curriculum Graph Co-Teaching (\ours) and thereafter Domain Curriculum Learning (\dcl) for the task of MTDA. We also discuss some preliminaries that are used to address the task.

\textbf{Problem Definition.} In the MTDA scenario, we are provided with a single source dataset $\mathcal{S} = \{(\mathbf{x}_{s,i}, y_{s,i})\}^{n_s}_{i=1}$, containing $n_s$ labeled samples, and $N$ unlabeled target datasets $\mathcal{T} = \{\mathcal{T}_j\}^{N}_{j=1}$, where $\mathcal{T}_j = \{\mathbf{x}_{t_j,k}\}^{n_j}_{k=1}$ with each containing $n_j$ unlabeled samples. As in any DA scenario, the fundamental assumption is that the underlying data distributions of the source and the targets are different from each other. It is also assumed that the label space of the source and targets are the same. Under these assumptions, the goal of the MTDA is to learn a single predictor for all the target domains by using the data in $\mathcal{S} \cup \{\mathcal{T}_j\}^{N}_{j=1}$.

\begin{figure*}[!t]
	\centering
	\includegraphics[width=0.95\linewidth]{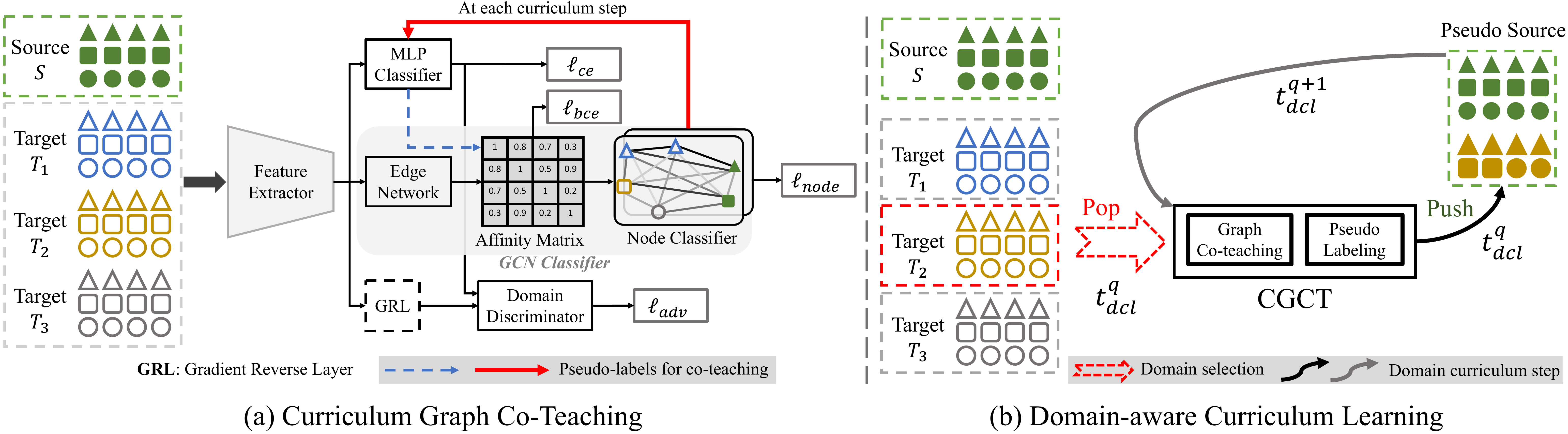}
	\vspace{-.12in}
	\caption{The pipeline of the proposed framework: a) \ours: Curriculum Graph Co-Teaching and b) DCL: Domain-aware curriculum learning. (a) In the \ours, the MLP Classifier provides pseudo-labels (PL) (\textcolor{blue}{$\dashrightarrow$} arrow) for the target samples to guide the Edge Network to learn the Affinity Matrix, whereas the Node Classifier of the GCN provides PL (bold \textcolor{red}{$\rightarrow$} arrow) to the MLP Classifier at the end of each curriculum step, realizing the \textit{co-teaching}. (b) In the DCL, the target domains are selected for adaptation, one at a time per domain curriculum step $t^{q}_{dcl}$, with the ``easier" domains selected first and then the ``harder" ones. After PL are obtained, the pseudo-labeled target dataset is added to the Pseudo Source dataset, which is then used in the next adaptation step.}
	\label{fig:flow}
	\vspace{-3mm}
\end{figure*}


\subsection{Preliminaries}

\textbf{Baseline for Multi-target Domain Adaptation.} Domain Adversarial Network (DANN)~\cite{ganin2016domain}, originally designed for STDA, aligns the feature distributions of the source and the target domains by using an adversarial training. DANN comprises of three networks: the feature extractor, the classifier and the domain discriminator. The classifier is responsible for classifying the features obtained from the feature extractor into $n_c$ classes. On the one hand, the domain discriminator distinguishes the source from the target features. While on the other hand, the feature extractor is trained to fool the discriminator and simultaneously learn good features for semantic classification.

Formally, let $F_\theta : \mathbb{R}^{3 \textrm{x} w \textrm{x} h} \rightarrow \mathbb{R}^d$ be the feature extractor network, parameterized by $\theta$, that outputs a feature $\mathbf{f} = F(\mathbf{x})$ for a given sample $\mathbf{x}$. The classifier network, parameterized by $\phi$, is denoted by $G_\phi : \mathbb{R}^{d} \rightarrow \mathbb{R}^{n_c}$, which takes as input a feature $\mathbf{f}$ and outputs class logits $\mathbf{g} = G(\mathbf{f})$. The discriminator network $D_\psi : \mathbb{R}^{d} \rightarrow \mathbb{R}^{1}$, parameterized by $\psi$, takes in the same feature $\mathbf{f}$ and outputs a single logit. By 
treating all the target domains as one combined target domain, the overall training objective of DANN for MTDA is given by:
\vspace{-3mm}
\begin{equation}
\label{eqn:dann}
    \Max_{\psi} \; \Min_{\theta, \phi} \; \ell_{ce} - \lambda_{adv} \: \ell_{adv},
\end{equation}
\vspace{-3mm}
\[
    \begin{split}
        \text{where} \; \ell_{ce} &= \displaystyle - \E_{(\mathbf{x}_{s,i}, y_{s,i}) \sim \mathcal{S}} \: \widetilde{y}_{s,i} \text{log} \: G(F(\mathbf{x}_{s,i})),\\
        \text{and}~\ell_{adv} & = \displaystyle - \E_{\mathbf{x}_{s,i} \sim \mathcal{S}} \: \text{log} \: D(F(\mathbf{x}_{s,i})) \\
                   & ~~~~~- \displaystyle \E_{x_{t, j} \sim \mathcal{T}} \: \text{log} \: [1 - D(F(\mathbf{x}_{t,j}))].
                   %
    \end{split}
\]
$\widetilde{y}_{s,i}$ is the one-hot label for a source label $y_{s,i}$. The first term, $\ell_{ce}$, in Eq.~\ref{eqn:dann} is the cross-entropy loss computed on the source domain samples and minimized w.r.t. $\theta, \phi$. The second term, $\ell_{adv}$, in Eq.~\ref{eqn:dann} is the adversarial loss that is maximized w.r.t $\psi$ but minimized w.r.t $\theta$. $\lambda_{adv}$ is the weighing factor for $\ell_{adv}$. To capture the multi-modal nature of the distributions, CDAN~\cite{long2018conditional} is proposed where $D$ can be additionally conditioned on the classifier predictions $\mathbf{g}$. In CDAN~\cite{long2018conditional}, the $D$ takes as input $\mathbf{h}=(\mathbf{f}, \mathbf{g})$, the joint variable of $\mathbf{f}$ and $\mathbf{g}$, instead of just $\mathbf{f}$. In this work we use CDAN for aligning the feature distributions.


\textbf{Graph Convolutional Network.} For the GCN~\cite{kipf2017semi} classifier we construct an undirected and fully-connected graph $\Gamma=(\mathcal{V}, \mathcal{E}, \mathcal{A})$ from all samples in mini-batch. In details, given a mini-batch of images, we represent each image $\mathbf{x}_i$ as a node $\mathbf{v}_i \in \mathcal{V}$ in the $\Gamma$. $e_{i, j} \in \mathcal{E}$ indicates an edge between nodes $\mathbf{v}_{i}$ and $\mathbf{v}_{j}$, and $a_{i,j}$ is the semantic similarity score for nodes $(\mathbf{v}_{i}$, $\mathbf{v}_{j})$ forming an affinity matrix $\mathcal{A}$.

Following~\cite{luo2020progressive}, we compute the semantic similarity scores $\hat{a}_{i,j}^{(l)}$ at the $l$-th layer for all pairs $(\mathbf{v}_{i}, \mathbf{v}_{j}) \in \mathcal{E}$:
\vspace{-3mm}
\begin{equation}
    \label{eqn:affinity}
    \hat{a}_{i,j}^{(l)} = f_{edge}^{(l)}(\mathbf{v}_{i}^{(l-1)}, \mathbf{v}_{j}^{(l-1)}),
    \vspace{-2mm}
\end{equation}
where $f_{edge}^{(l)}$ is a non-linear similarity function parameterized by $\varphi$, and $\mathbf{v}_{i}^{(l-1)}$ is features at $l$-1 GCN layer of a sample $\mathbf{v}_{i}$. The initial node features $\mathbf{v}_{i}$ are instantiated with $\mathbf{f}_i$, the embedding obtained from $F$.
Then, we add self-connections for nodes in the graph and normalize the obtained similarity scores as:

\vspace{-1mm}
\begin{equation}
    \mathcal{A}^{(l)} = M^{-\frac{1}{2}}(\hat{\mathcal{A}}^{(l)} + I)M^{-\frac{1}{2}},
\end{equation}
where $M$ is the degree matrix, $I$ is the identity matrix, and $\hat{\mathcal{A}}$ is the un-normalized affinity matrix.

Finally, given the affinity matrix $\mathcal{A}^{(l)}$, we update the node features with the following propagation rule:
\begin{equation}
    \label{eqn:node_classifier}
    \mathbf{v}_{i}^{(l)} = f_{node}^{(l)} \Big([\mathbf{v}_{i}^{(l-1)}, \sum_{j \in \mathcal{B}} a_{i,j}^{(l)}\cdot \mathbf{v}_{j}^{(l-1)}]\Big),
    \vspace{-2mm}
\end{equation}
where $f_{node}^{(l)}$ is a non-linear function parameterized by $\varphi'$, $\mathcal{B}$ is a set of samples in the mini-batch, and $[\cdot, \cdot]$ is the feature concatenation function. The final $f_{node}^{(L)}$ layer is the output layer with $n_c$ outputs. We slightly abuse the notations and drop the superscript $l$ in our subsequent formulations for the sake of clarity.

\vspace{-3mm}
\subsection{Curriculum Graph Co-Teaching}
\label{sec:cgct}
\vspace{-3mm}
In this work we introduce the Curriculum Graph Co-Teaching (\ours) that employs feature aggregation with a GCN and uses curriculum learning for pseudo-labeling. In details, as shown in Fig.~\ref{fig:flow}(a), it is composed of: a feature extractor $F$, a domain discriminator $D$, a MLP classifier $G_{mlp}$ and a GCN classifier $G_{gcn}$.  The $G_{mlp}$ is a fully-connected output layer with $n_c$ output logits. The $G_{gcn}$ consists of an edge network $f_{edge}$ and a node classifier $f_{node}$. The $f_{node}$ aggregates the features of the samples in $\mathcal{B}$ by considering the learnt pairwise similarity in the affinity matrix $\hat{\mathcal{A}}$ produced by the  $f_{edge}$. The $G_{gcn}$ also outputs $n_c$ logits. Since, the $G_{mlp}$ and the $G_{gcn}$ capture different aspects of learning, they are exploited to provide feedback to each other in a co-teaching fashion. The \ours is trained for $Q$ \textit{curriculum steps} where a curriculum step, $t^{q}_{cur}$, is an episode in which the network is trained for $K$ training iterations. Each curriculum step $t^{q}_{cur}$ is further decomposed into two stages: i) Adaptation stage and ii) Pseudo-labeling stage. Each stage in a $t^{q}_{cur}$ is described below. Note that, as in~\cite{chen2019blending}, we assume that the domains labels of the target are latent and not observed during training.

\textbf{Adaptation stage}. In this stage we mainly perform the feature alignment using CDAN~\cite{long2018conditional}. In details, initially at step $t^{0}_{cur}$ we start with a source set $\hat{\mathcal{S}}^{0} = \{\mathcal{S}\}$ and a target set $\mathcal{T}$. We sample mini-batches $\mathcal{B}^{0} = \{\mathcal{B}^{0}_{s}, \mathcal{B}^{0}_{t}\} = \{\mathcal{B}^{0}_{s, i}, \mathcal{B}^{0}_{t, i}\}^{B}_{i=1}$ with size $B$ such that $\mathcal{B}^{0}_{s, i} \sim \hat{\mathcal{S}}$ and $\mathcal{B}^{0}_{t, i} \sim \mathcal{T}$. Each mini-batch of images is first fed to the feature extractor $F$ to obtain $\mathcal{F}^{0} = \{\mathbf{f}^{0}_{s,i}, \mathbf{f}^{0}_{t,i}\}^{B}_{i=1}$ which are then simultaneously fed to both the $G_{mlp}$ and $G_{gcn}$. When fed to the $G_{mlp}$ it outputs the logits $\hat{\mathcal{G}}^{0} = \{\hat{\mathbf{g}}^{0}_{s,i}, \hat{\mathbf{g}}^{0}_{t,i}\}^{B}_{i=1}$. On the other hand, $\mathcal{F}^{0}$ are input to the $f_{edge}$ to estimate the pairwise similarity of the samples in $\mathcal{B}^{0}$. Specifically, the $f_{edge}$ outputs an affinity matrix $\hat{\mathcal{A}}$ following Eq.~\ref{eqn:affinity}, where the entries $\hat{a}_{i,j}$ in $\hat{\mathcal{A}}$ denote the strength of similarity between samples $i$ and $j$ in $\mathcal{B}^{0}$. Intuitively, higher the value of $\hat{a}_{i,j}$, higher is the likelihood of samples $i$ and $j$ belonging to the same semantic category. Finally, following Eq.~\ref{eqn:node_classifier}, the $f_{node}$ aggregates the features in $\mathcal{F}^{0}$ based on the estimated $\hat{\mathcal{A}}$ such that for each node the most similar samples in the neighbourhood contribute more to its final representation. Subsequently, the $f_{node}$ outputs its logits as $\bar{\mathcal{G}}^{0} = \{\bar{\mathbf{g}}^{0}_{s,i}, \bar{\mathbf{g}}^{0}_{t,i}\}^{B}_{i=1}$. The elements in $\hat{\mathcal{G}}^{0}$ and $\bar{\mathcal{G}}^{0}$ are then passed through a softmax function to obtain the probabilities for each sample as $p(\hat{y}=c|\hat{\mathbf{g}}; c \in n_c)$ and $p(\bar{y}=c|\bar{\mathbf{g}}; c \in n_c)$, where $\hat{y}$ and $\bar{y}$ are the predictions, respectively.

To guide the $f_{edge}$ to learn the pairwise similarity between the samples in $\mathcal{B}^{0}$ we propose the concept of co-teaching where the $G_{mlp}$ provides feedback to the $f_{edge}$. Since, $G_{mlp}$ makes instance-level independent predictions on the samples in $\mathcal{B}^{0}$, it is not susceptible to the accumulation of potential noise from the dissimilar neighbours. To this end, for a $\mathcal{B}^{0}$ we construct a ``\textit{target}" affinity matrix $\hat{\mathcal{A}}^{tar}$ and enforce the predictions of $f_{edge}$ to be as close as possible to the $\hat{\mathcal{A}}^{tar}$. Each entry $\hat{a}^{tar}_{i,j}$ in the $\hat{\mathcal{A}}_{tar}$ is given by:
\vspace{-2mm}
\begin{equation}
\vspace{-2mm}
\label{eqn:tar_affinity}
    \hat{a}^{tar}_{i,j} = 
    \begin{cases}
        1, \;\;\;\text{if } y_i = y_j = c\\
        0, \;\;\; \text{otherwise}
    \end{cases},
\end{equation}
where $c$ is the class label. While the class labels of $\mathcal{B}^{0}_{s}$ are provided as ground truth, we do not have access to the labels of $\mathcal{B}^{0}_{t}$. Therefore, a target domain sample $\mathbf{x}_{t,j} \in \mathcal{B}^{0}_{t}$ is assigned a definitive pseudo-label $\hat{y}_{t,j} = c'$ where $c' = \ArgMax_{c \in n_c} p(\hat{y}_{t,j} = c|\hat{\mathbf{g}}_{t,j})$ if the maximum likelihood $\Max_{c \in n_c}p(\hat{y}_{t,j} = c|\hat{\mathbf{g}}_{t,j})$ is greater than a threshold $\tau$. The entries $\hat{a}^{tar}_{i,j}$ involving $\mathbf{x}_{t,j} \in \mathcal{B}^{0}_{t}$ not passing the $\tau$ are not optimized during training. We train the $f_{edge}$ using a binary cross-entropy loss as:
\begin{equation}
\label{eqn:bce}
    \ell^{edge}_{bce} = \hat{a}^{tar}_{i,j} \; \text{log } p(\hat{a}_{i,j}) + (1 - \hat{a}^{tar}_{i,j}) \; \text{log }(1 - p(\hat{a}_{i,j})).
\end{equation}

Finally, for training the $G_{mlp}$ and the $f_{node}$ in the $G_{gcn}$ we compute the standard cross-entropy loss with the samples in $\mathcal{B}^{0}_{s}$ as:

\vspace{-3mm}
\begin{equation}
\label{eqn:ce_mlp}
\ell^{mlp}_{ce} = - \frac{1}{|\mathcal{B}^{0}_{s}|} \displaystyle \sum^{|\mathcal{B}^{0}_{s}|}_{i=1} \Tilde{y}_i \; \text{log } \; p(\hat{y}_{s,i}|\hat{\mathbf{g}}^{0}_{s,i}),
\end{equation}
\vspace{-2mm}
\begin{equation}
\label{eqn:ce_mlp}
\ell^{node}_{ce} = - \frac{1}{|\mathcal{B}^{0}_{s}|} \displaystyle \sum^{|\mathcal{B}^{0}_{s}|}_{i=1} \Tilde{y}_i \; \text{log } \; p(\bar{y}_{s,i}|\bar{\mathbf{g}}^{0}_{s,i}).
\end{equation}

We feed the features $\{\hat{\mathbf{h}}^{0}_{s,i}, \hat{\mathbf{h}}^{0}_{t,i}\}^{B}_{i=1} = \{(\mathbf{f}^{0}_{s,i}, \hat{\mathbf{g}}^{0}_{s,i}), (\mathbf{f}^{0}_{t,i}, \hat{\mathbf{g}}^{0}_{t,i})\}^{B}_{i=1}$, corresponding to $\mathcal{B}^{0}$, to the domain discriminator $D$ and compute the conditional adversarial loss following Eq.~\ref{eqn:dann}. Thus, the final objective function for the \ours can be written as:
\vspace{-2mm}
\begin{equation}
\label{eqn:cgct_loss}
    \begin{split}
        \max_{\psi} \; \min_{\theta, \phi, \varphi, \varphi'} \; \ell^{mlp}_{ce} &+ \lambda_{edge}\ell^{edge}_{bce} \\
        &+ \lambda_{node}\ell^{node}_{ce} - \lambda_{adv} \ell_{adv},
    \end{split}
\end{equation}
where $\lambda_{edge}$, $\lambda_{node}$ and $\lambda_{adv}$ are the weighing factors. 

\textbf{Pseudo-labelling stage.} Upon completion of the adaptation stage in a curriculum step $t^{q}_{cur}$ we put the network in inference mode and obtain pseudo-labels $\forall \mathbf{x}_{t,j} \in \mathcal{T}$. The $G_{gcn}$ is employed for this task because, owing to its aggregating characteristics, it learns more robust features~\cite{yang2020heterogeneous} than the $G_{mlp}$. This is the \textit{curriculum} aspect of our proposed co-teaching training strategy in \ours where the obtained pseudo-labeled target samples are then used to train the $G_{mlp}$, besides the $f_{node}$.

At any step $t^{q}_{cur}$, the criterion for pseudo-label selection is formally written as:
\vspace{-2mm}
\begin{equation}
\label{eqn:pseudo_label_stage}
\forall \mathbf{x}_{t,j} \in \mathcal{T}, w_{j} = 
\begin{cases}
    1, \; \; \; \text{if }  \Max_{c \in n_c}p(\bar{y}_{t,j} = c|\bar{\mathbf{g}}_{t,j}) > \tau \\
    0, \; \; \; \text{otherwise }
\end{cases},
\vspace{-1.5mm}
\end{equation}
where $w_j = 1$ signifies that $\mathbf{x}_{t,j}$ is selected with a pseudo-label $\bar{y}_{t,j} = c'$ where $c' = \ArgMax_{c \in n_c} p(\bar{y}_{t,j} = c|\bar{\mathbf{g}}_{t,j})$, whereas $w_j = 0$ denotes no pseudo-label is assigned. After the pseudo-labeling stage in a $t^{q}_{cur}$ we obtain a pseudo-labeled target set $\mathcal{D}^{q}_{t} = \{(\mathbf{x}_{t,j}, \bar{y}_{t,j})\}^{\bar{n}_t}_{j=1}$ where $\bar{n}_t$ is the number of recruited pseudo-labeled target samples. Post pseudo-labeling we update and prepare the source set for the succeeding step $t^{q+1}_{cur}$ as:
\vspace{-1.5mm}
\begin{equation}
\label{eqn:set_update}
\hat{\mathcal{S}}^{q+1} = \mathcal{S} \; \cup \; \mathcal{D}^{q}_{t}.
\vspace{-2mm}
\end{equation}

\noindent The update rule in Eq.~\ref{eqn:set_update} allows us to compute the supervised losses $\ell^{node}_{ce}$ and $\ell^{mlp}_{ce}$ from Eq.~\ref{eqn:cgct_loss} for $\mathbf{x}_{t,j} \sim \mathcal{D}_{t}$. Note that we do not alter the domain labels in $\mathcal{D}^{q}_{t}$ and hence, the formulation for $\ell_{adv}$ remains unchanged. 

At the culmination of $Q$ curriculum steps, $\hat{\mathcal{S}}^{Q}$ is obtained using Eq.~\ref{eqn:set_update} and the network is fine-tuned with only the supervised losses in Eq.~\ref{eqn:cgct_loss} for $K'$ training iterations.

\subsection{Domain-aware Curriculum Learning}
\label{sec:domain_curr}
Now we consider the case when the domain labels of the target are available, \ie $\mathcal{T} = \{\mathcal{T}_j\}^{N}_{j=1}$, $N$ being the number of target domains. In principle, when the domain labels are available, one can either train $N$ domain discriminators or a ($N+1$) way single domain discriminator. Apart from over-parameterization, it also suffers from limited gradients coming from the discriminator(s) due to single point estimates~\cite{kurmi2019curriculum}. Thus, we propose \textbf{D}omain-aware \textbf{C}urriculum \textbf{L}earning (DCL) as an alternate learning paradigm to better utilize the target domain labels in the MTDA setting.

To this end we design the DCL that is based on our proposed Easy-to-Hard Domain Selection (EHDS) strategy. Our proposal for the DCL stems from the observation that different target domains exhibit different domain shifts from the source domain, where some domain shifts are larger than the others. Evidently, the network will find it easier to adapt to the closest target domain while performing sub-optimally on the domain with the largest domain shift. When adaptation is performed with $N$ domains at tandem then the large domain shifts of harder domains will interfere with the feature alignment on the easier target domains, thereby compromising the overall performance. To overcome this problem, in the EHDS strategy, as the name suggests, the network performs feature adaptation one domain at a time, starting from the easiest target domain and gradually moving towards the hardest. The ``\textit{closeness}" of a target domain from the source is measured by the uncertainty in the target predictions with a source-trained model. Lesser the uncertainty in predictions, closer the target from the source domain. Therefore, measuring the entropy on a target domain can serve as a good proxy for domain selection, and is defined as:
\vspace{-3mm}
\begin{equation}
\label{eqn:domain_metric}
H(\mathcal{T}_j) = - \under\limits_{\mathbf{x}_{t_j,k} \sim \mathcal{T}_j} \displaystyle \sum^{|n_c|}_{c=1} p(\hat{y}_{t_j,k,c}|\mathbf{x}_{t_j,k}) \text{log }p(\hat{y}_{t_j,k,c}|\mathbf{x}_{t_j,k}).
\vspace{-1mm}
\end{equation}

Due to this step-by-step adaptation through domain traversal, the intermediate target domains help in reducing large domain shifts by making the farthest domain shift considerably closer than that at the start. Differently from the \ours, in the DCL, each curriculum step, defined as $t^{q}_{dcl}$, consist in learning over one target domain, with a total of $N$ steps. Since, the simulation of single-source and single-target adaptation inside the MTDA setup yields better domain-invariant features, at the end of each $t^{q}_{dcl}$ we also consider extracting pseudo-labels for the target samples from the classifier and add them to the source set (see Fig.~\ref{fig:flow}(b)) for computing the supervised losses. This further reduces the domain gaps for the forthcoming harder domains. The $t^{q}_{dcl}$ is split into three stages and are described below:

\textbf{Domain selection stage.} Given a source-trained model $F_{\theta^{*}}(G_{\phi^{*}})$, where $\theta^{*}$ and $\phi^{*}$ are the trained parameters of $F$ and $G$, and initial source and target sets $\hat{\mathcal{S}}^{0} = \{\mathcal{S}\}$ and $\hat{\mathcal{T}}^{0} = \{\mathcal{T}_j\}^{N}_{j=1}$, the closest target domain is selected as:
\vspace{-2mm}
\begin{equation}
\label{eqn:domain_transferability}
\mathbb{D}^{0} = \Argmin_{j} \{H_j (\mathcal{T}_{j}) \; | \; \forall \mathcal{T}_j \in \hat{\mathcal{T}}^{0}\},
\vspace{-2mm}
\end{equation}

\noindent where $\mathbb{D}^{0}$ is the target domain selected at step $t^{0}_{dcl}$ and is used for performing adaptation in the subsequent stage.

\textbf{Adaptation stage.} This stage is similar to the one in $t^{q}_{cur}$, described in Sec.~\ref{sec:cgct}, except the feature adaptation at any step $t^{q}_{dcl}$ is performed using $\hat{\mathcal{S}}^{q} \cup \mathcal{T}_{\mathbb{D}^{q}}$, rather than the entire target set $\mathcal{T}$. The model is trained using the losses described in Eq.~\ref{eqn:cgct_loss}.

\textbf{Pseudo-labeling stage.} The criterion for pseudo-label selection still remains the same, as described in Eq.~\ref{eqn:pseudo_label_stage}, with the exception of target samples being drawn only from the current target domain $\mathbb{D}^{q}$, yielding a pseudo-labeled target set $\mathcal{D}^{\mathbb{D}^{q}}_{t}$. Consequently, the source and target set update changes as following:
\vspace{-2mm}
\begin{equation}
\label{eqn:dcl_source_update}
\hat{\mathcal{S}}^{q+1} = \hat{\mathcal{S}}^{q} \; \cup \; \mathcal{D}^{\mathbb{D}^{q}}_{t},
\vspace{-2mm}
\end{equation}

\vspace{-2mm}
\begin{equation}
\label{eqn:dcl_tgt_update}
\hat{\mathcal{T}}^{q+1} = \hat{\mathcal{T}}^{q} \; \text{\textbackslash } \; \mathcal{T}_{\mathbb{D}^{q}}.
\vspace{-2mm}
\end{equation}

These three stages are repeated until all $N$ domains have been exhausted. Then similarly, as in \ours, the final model is fine-tuned with $\hat{\mathcal{S}}^{Q}$. When \ours is trained using the DCL strategy we refer to the model as \domours. We would like to point that the DCL can also be realized with a single classifier model (see Sec.~\ref{sec_exp}).


%% file: experiments.tex
\vspace{-3mm}
\section{Experiments}
\label{sec_exp}
\subsection{Dataset and Experimental Details}

\textbf{Datasets.} 
We conduct experiments on five standard UDA benchmarks: Digits-five~\cite{xu2018deep}, Office-31~\cite{saenko2010adapting}, PACS~\cite{li2017deeper}, Office-Home~\cite{venkateswara2017Deep} and the very large scale DomainNet~\cite{peng2019moment} (\textbf{0.6 million} images). The statistics of the datasets are summarized  in Tab.~\ref{tab:mtda_datasets}. More details on the datasets can be found in the Supp. Mat.


\begin{table}[!h]
    \centering
    \small
    \def\arraystretch{.9}
    \setlength{\tabcolsep}{5.5pt}
    \begin{tabular}{lccc}
         \hline
         \thead{Dataset}& \thead{\#domains} & \thead{\#classes} & \thead{\#images} \\
         \hline
         Digits-five & 5 & 10 & $\sim$ 145K \\
         PACS & 4 & 7 & 9,991 \\
         Office-31 & 3 & 31 & 4,652 \\
         Office-Home & 4 & 65 & 15,500 \\
         DomainNet & 6 & 345 & $\sim$ 0.6M\\
         \hline
    \end{tabular}
    \vspace{-.12in}
    \caption{Dataset details for multi-target domain adaptation.}
    \label{tab:mtda_datasets}
    \vspace{-3mm}
\end{table}





\textbf{Evaluation protocol.} We use the classification accuracy to evaluate the performance. The classification accuracy is computed for every possible combination of one source domain and the rest of the target domains. The performance for a given direction, \ie, \textit{source}$\rightarrow$\textit{rest}, is given by averaging the accuracy on all the target domains, where \textit{source} signifies the source domain and \textit{rest} indicates all the unlabeled domains except the \textit{source}. Importantly, in all our experiments we always report the final classification accuracy obtained with the $G_{mlp}$ because the $G_{gcn}$ always requires a mini-batch at inference, an assumption which is easily violated when deployed in the real world.

\textbf{Implementation details.} To be fairly comparable with the state-of-the-art methods, we adopt the backbone feature extractor networks used in~\cite{chen2019blending, yang2020heterogeneous, peng2019domain} for the corresponding datasets. We train the networks by using a Stochastic Gradient Descent (SGD) optimizer having an initial learning rate of 1e-3 and decay exponentially. More details about the network architecture and experimental set-up can be found in the Supp. Mat. 

\textbf{Hyperparameter selection.} In our final model we used only a single set of hyperparameters, which are $\lambda_{edge}=1$, $\lambda_{node}=0.3$, $\lambda_{adv}=1$ and $\tau = 0.7$. Following the standard protocol in~\cite{shu2018dirt}, we used a held-out validation set of 1000 samples for the MNIST $\rightarrow$ \textit{rest} direction to tune these hyper-parameters.

\vspace{-1mm}
\subsection{Ablations}
\label{sec:ablations}

In this section we discuss the design choices of our proposed contributions and report the results of a thorough ablation study. Our ablation analysis highlights the importance of the \textit{graph co-teaching} and the \textit{curriculum learning}. We run the ablation experiments on Office-Home with ResNet-18~\cite{he2016deep} as backbone network and on Digits-five with a network adopted from AMEAN~\cite{chen2019blending}. We adopt the CDAN as a baseline for adaptation in Tab.~\ref{tab:coteaching-ablations} and Tab.~\ref{tab:abl_table}.

\begin{table}[!h]
    \centering
    \small
    \setlength{\tabcolsep}{3.pt}
    \begin{tabular}{lcccc|c}
    \specialrule{1.5pt}{1pt}{1pt} 
    & & \multicolumn{3}{c|}{Pseudo-labels from}&\\
    \cline{3-5}
    Model & Co-teaching & $G_{mlp}$ & $f_{edge}$ & $f_{node}$ & \textbf{Avg}(\%) \\
    \specialrule{1.5pt}{1.pt}{1pt}
    M1 & \xmark & self & $G_{mlp}$ & $G_{mlp}$ & 57.4 \\
    \rowcolor{Gray}
    M2 & \xmark & $G_{gcn}$ & $G_{gcn}$ & $G_{gcn}$ & 59.6 \\
    M3 & \cmark & self & \makecell{$G_{mlp}$, \\$G_{gcn}$} & $G_{mlp}$ & 58.2 \\
    \rowcolor{Gray}
    \domours (Ours) & \cmark & $G_{gcn}$ & $G_{mlp}$ & $G_{gcn}$ & \textbf{60.8} \\
    \specialrule{1.5pt}{1pt}{1pt}
    \end{tabular}
    \vspace{-.12in}
    \caption{Ablation study of different co-teaching strategies on Office-Home. We reported the classification accuracy averaged across all the \textit{source} $\rightarrow$ \textit{rest} directions.}
    \label{tab:coteaching-ablations}
    \vspace{-3mm}
\end{table}

\textbf{Graph co-teaching.} The goal of this particular ablation study is to analyse why our proposed graph co-teaching is beneficial and the manner in which it should be realised in an adaptation framework. To this end, as shown in the Tab.~\ref{tab:coteaching-ablations}, we design some baselines that can be distinguished in the manner in which the $G_{mlp}$ and the $G_{gcn}$ provide pseudo-labels to the each other (columns 3 to 5) and then compare it to our \domours. In more details, the baseline models can be described as: i) M1: a baseline where the $G_{mlp}$ provides pseudo-labels to itself, $f_{edge}$ and $f_{node}$ after each curriculum step  $t^{q}_{dcl}$; ii) M2: a baseline similar to M1, except that the $G_{gcn}$ provides the pseudo-labels; iii) M3: another baseline which is similar to M1 but with an exception that the $G_{gcn}$ also provides pseudo-labels to $f_{edge}$ for the current target domain in an ongoing $t^{q}_{dcl}$ step.

\begin{table*}[!h]
    \centering
    \small
    \def\arraystretch{.9}
    \setlength{\tabcolsep}{6.0pt}
    \begin{tabular}{l|lcccccccccc}
    \specialrule{1.5pt}{1pt}{1pt}
    && \multicolumn{4}{c}{Office-31} & &  \multicolumn{5}{c}{Office-Home}\\
    \cline{3-6} \cline{8-12}
    Setting&Model & Amazon &DSLR & Webcam & \textbf{Avg}(\%) & & Art & Clipart & Product & Real & \textbf{Avg}(\%) \\
    \specialrule{1.5pt}{1pt}{1pt}
    \makecell{w/o Target}&Source train & 68.6 & 70.0 & 66.5 & 68.4 & & 47.6 & 42.6 & 44.2 & 51.3 & 46.4\\
    \specialrule{0.5pt}{1pt}{1pt}
    \multirow{5}{*}{\makecell{Single-\\Target}}&DAN~\cite{long2015learning} & 79.5 & 80.3 & 81.2 & 80.4 && 56.1 & 54.2 &51.7&63.0&56.3\\
    &RevGrad~\cite{ganin2016domain} & 80.8 & 82.5 & 83.2 & 82.2 &&58.3&55.4&52.8&63.9&57.6\\
    &JAN~\cite{long2016deep} & 85.0 & 83.0 & 85.6 & 84.3 && 58.7& 57.0&53.1&64.3&58.3\\
    &CDAN~\cite{long2018conditional} & \textbf{91.4} & 84.1 & 84.0 & 86.6 && 64.2&62.9&59.9&68.1&63.8\\
    
    & \textbf{CGCT} (ours)
    & 89.6 &\textbf{85.5}&\textbf{87.6}&\textbf{87.6}& &\textbf{67.9}&\textbf{68.7}& \textbf{62.3}&\textbf{70.7} & \textbf{67.4}\\
    \specialrule{0.5pt}{1pt}{1pt}
    \multirow{6}{*}{\makecell{Target-\\Combined}}&DAN~\cite{long2015learning} & 78.0 & 64.4 & 66.7 & 69.7 & & 55.6 & 56.6 & 48.5 & 56.7 & 54.4 \\
    
    &RevGrad~\cite{ganin2016domain} & 78.2 & 72.2 & 69.8 & 73.4 & & 58.4 & 58.1 & 52.9 & 62.1 & 57.9\\
    &JAN~\cite{long2016deep} & 84.2 & 74.4 & 72.0 & 76.9 & & 58.3 & 60.5 & 52.2 & 57.5 & 57.1\\
    &CDAN~\cite{long2018conditional} & 93.6	& 80.5	& 81.3	& 85.1 & & 59.5 & 61.0 & 54.7	& 62.9& 59.5\\
    &AMEAN~\cite{chen2019blending} & 90.1 & 77.0 & 73.4 & 80.2 & & 64.3 & 65.5 & 59.5 & 66.7 & 64.0\\

    &\textbf{CGCT} (ours) &\textbf{93.9}&\textbf{85.1}& \textbf{85.6}&\textbf{88.2}& &\textbf{67.4}&\textbf{68.1}&\textbf{61.6} &\textbf{68.7} &\textbf{66.5}\\
    \specialrule{0.5pt}{1pt}{1pt}
    
    \multirow{4}{*}{\makecell{Multi-\\Target}}&MT-MTDA~\cite{nguyen2020unsupervised} & 87.9 & 83.7 & 84.0 & 85.2 & & 64.6 & 66.4 & 59.2 & 67.1 & 64.3\\
    &HGAN~\cite{yang2020heterogeneous} & 88.0 & 84.4 & 84.9 & 85.8 & & - & - & - & - & -\\
    &\textbf{CDAN+DCL} (ours) & 92.6 & 82.5 & 84.7 & 86.6  & & 63.0 & 66.3 & 60.0 & 67.0 & 64.1 \\
    
    &\textbf{D-CGCT} (ours) &\textbf{93.4} &\textbf{86.0} &\textbf{87.1} &\textbf{88.8} &&\textbf{70.5}&\textbf{71.6}&\textbf{66.0}&\textbf{71.2}&\textbf{69.8}\\
    
    \specialrule{1.5pt}{1pt}{1pt}
    \end{tabular}
    \vspace{-.12in}
    \caption{Comparison with state-of-the-art methods on Office-31 and Office-Home. All methods use the ResNet-50 as the backbone.
    Single-Target indicates methods are performed on one source to one target setting. Target-Combined indicates methods are performed on one source to aggregated targets setting, while the Multi-Target indicates methods are performed on one source to multi-target setting.}
    \label{tab:SOTA-Office31-Home}
    \vspace{-5mm}
\end{table*}

Unsurprisingly, M1 performs the worst of all the baselines because the pseudo-labels computed by the $G_{mlp}$ are less accurate due to $G_{mlp}$ not taking into account the \textit{feature aggregation} from multiple domains. Contrarily, the baseline M2 performs better than the M1 due to the fact that M2 uses $G_{gcn}$ for pseudo-labeling, which are more accurate. This highlights the importance of feature aggregation in the MTDA setting. One other thing that separates \domours from both M1 and M2 is the co-teaching, which is absent in the latter baselines. Since, the \domours enables co-teaching, with the $G_{mlp}$ and the $G_{gcn}$ providing pseudo-labels to each other, it does not overfit on the same ``incorrect" pseudo-label, thereby achieving more robust predictions. Contrarily, M3 uses co-teaching and yet it fails to achieve comparable performance. We speculate that, since the $f_{edge}$ is also trained with the pseudo-labels obtained from the $G_{gcn}$ for the current target domain in a $t^{q}_{dcl}$ step, it becomes susceptible to noise. Thus, in summary, the graph co-teaching is the most effective when the $G_{gcn}$ is exploited to provide pseudo-labels only after each curriculum step.

\begin{table}[!h]
    \centering
    \small
    \def\arraystretch{.9}
    \setlength{\tabcolsep}{3.2pt}
    \begin{tabular}{lccccc}
    \specialrule{1.5pt}{1pt}{1pt}
    & \multicolumn{5}{c}{Office-Home}\\
    \cline{2-6}
    Model & Art & Clipart & Product & Real & \textbf{Avg}(\%) \\
    \specialrule{1.5pt}{1pt}{1pt}
    \rowcolor{Gray}
    Source train &  51.45 & 43.93 & 42.41 & 54.50 & 48.07\\
    Baseline & 50.70 & 50.78 & 47.95 & 57.63 & 51.77 \\
    \rowcolor{Gray}
    Base.\textsuperscript{$\dagger$} & 52.08 & 53.21 & 48.62 & 58.49 & 53.10 \\
    Base.\textsuperscript{$\dagger$}+PL & 54.61 & 56.13 & 50.25 & \textbf{61.04} & 55.51 \\
    \rowcolor{Gray}
    \textbf{Base.}\textsuperscript{$\dagger$} + \textbf{DCL} & \textbf{55.94} & \textbf{56.66} & \textbf{52.85} & 60.18 & \textbf{56.41}\\
    \specialrule{0.5pt}{1pt}{1pt}
    Base.\textsuperscript{$\dagger$}+GCN$\ddagger$ & 50.19 & 49.09 & 46.52 & 60.76 & 51.64 \\
    \rowcolor{Gray}
    Base.\textsuperscript{$\dagger$}+GCN$\ddagger$ + PL & 54.52 & 57.60 & 53.20 & 65.49 & 57.70 \\
    
    \textbf{\ours} & 60.81 & 60.00 & 54.13 & 62.62 & 59.39\\
    \rowcolor{Gray}
    
    \textbf{\domours} & \textbf{61.42} & \textbf{60.73} & \textbf{57.27} & \textbf{63.8} & \textbf{60.81}\\
    
    \specialrule{1.5pt}{1pt}{1pt}
    \end{tabular}
    \vspace{-.12in}
    \caption{Ablation results of different baselines using ResNet-18 as backbone on Office-Home. 
    \textbf{Baseline}: CDAN~\cite{long2018conditional} model that combines all the target domains into a single target domain. ``$\dagger$" indicates the baseline models that use the domain labels of the target. \textbf{GCN}$\ddagger$: the baseline model with the GCN as the single classification head. \textbf{PL}: using pseudo-labels.}
    \label{tab:abl_table}
    \vspace{-5mm}
\end{table}

\begin{table}[!h]
    \centering
     \small
     \def\arraystretch{.9}
     \setlength{\tabcolsep}{3.6pt}
    \begin{tabular}{Lccccccc}
    \specialrule{1.5pt}{1pt}{1pt}
    & \multicolumn{7}{c}{DomainNet}\\
    \cline{2-8}
    Model & Cli.	& Inf.	& Pai. & Qui. & Rea. & Ske. & \textbf{Avg}(\%)\\
    \specialrule{1.5pt}{1pt}{1pt}
    Source train & 25.6 & 16.8&25.8&9.2&20.6&22.3&20.1 \\
    SE~\cite{french2018self} & 21.3&8.5&14.5&13.8&16.0&19.7&15.6 \\
    MCD~\cite{saito2018maximum} & 25.1&19.1&27.0&10.4&20.2&22.5&20.7 \\
    DADA~\cite{peng2019domain} & 26.1&20.0& 26.5&12.9&20.7&22.8&21.5 \\
    CDAN~\cite{long2018conditional} & 31.6&27.1&31.8&12.5&33.2&35.8&28.7 \\
    MCC~\cite{jin2020minimum}	& 33.6&30.0&32.4&13.5&28.0& 35.3& 28.8\\
    \specialrule{0.5pt}{1pt}{1pt}
    \bf CDAN + DCL & 35.1 & 31.4 & 37.0 & \textbf{20.5} & 35.4 & \textbf{41.0} & 33.4\\
    \bf CGCT & 36.1&\textbf{33.3}&35.0&10.0&39.6&39.7&32.3\\
    
    \bf D-CGCT & \textbf{37.0}&32.2&\textbf{37.3}&19.3&\textbf{39.8}&40.8&\textbf{34.4}\\
    
    \specialrule{1.5pt}{1pt}{1pt}
    \end{tabular}
    \vspace{-.12in}
    \caption{Comparison with the state-of-the-art methods on DomainNet. All methods use the ResNet-101 as the backbone. The classification accuracy are reported for each \textit{source}$\rightarrow$\textit{rest} direction, with each \textit{source} domain being indicated in the columns. All the reported numbers are evaluated on the multi-target setting.}
    \label{tab:SOTA-DomainNet}
    \vspace{-5mm}
\end{table}

\textbf{Curriculum learning.} We also study the effect of domain-aware curriculum learning in isolation from co-teaching. For that purpose, as shown in the Tab.~\ref{tab:abl_table}, we start with the baseline model CDAN by treating all the target domains as one single domain. When the domain labels of the target are available, the baseline improves by 1.33\%, indicating that the domain labels can indeed improve the performance of an adaptation model. To show the benefit of the DCL without co-teaching, we train the \textbf{Base}\textsuperscript{$\dagger$} + \textbf{DCL}, and it yields an average accuracy that is higher than the Base.\textsuperscript{$\dagger$} + PL counterpart. The advantage of using DCL is further amplified when coupled with the \ours, where the \domours outperforms all other baselines, including the \ours. Due to the gradual adaptation, the \domours also leads to the better cluster formation than the \ours, as shown by the \textit{t}-SNE visualization in the Fig.~\ref{tab:tsne}.

\begin{table}[!h]
    \setlength{\tabcolsep}{.1pt}
    \begin{tabular}{ll}
        \includegraphics[width=.5\linewidth]{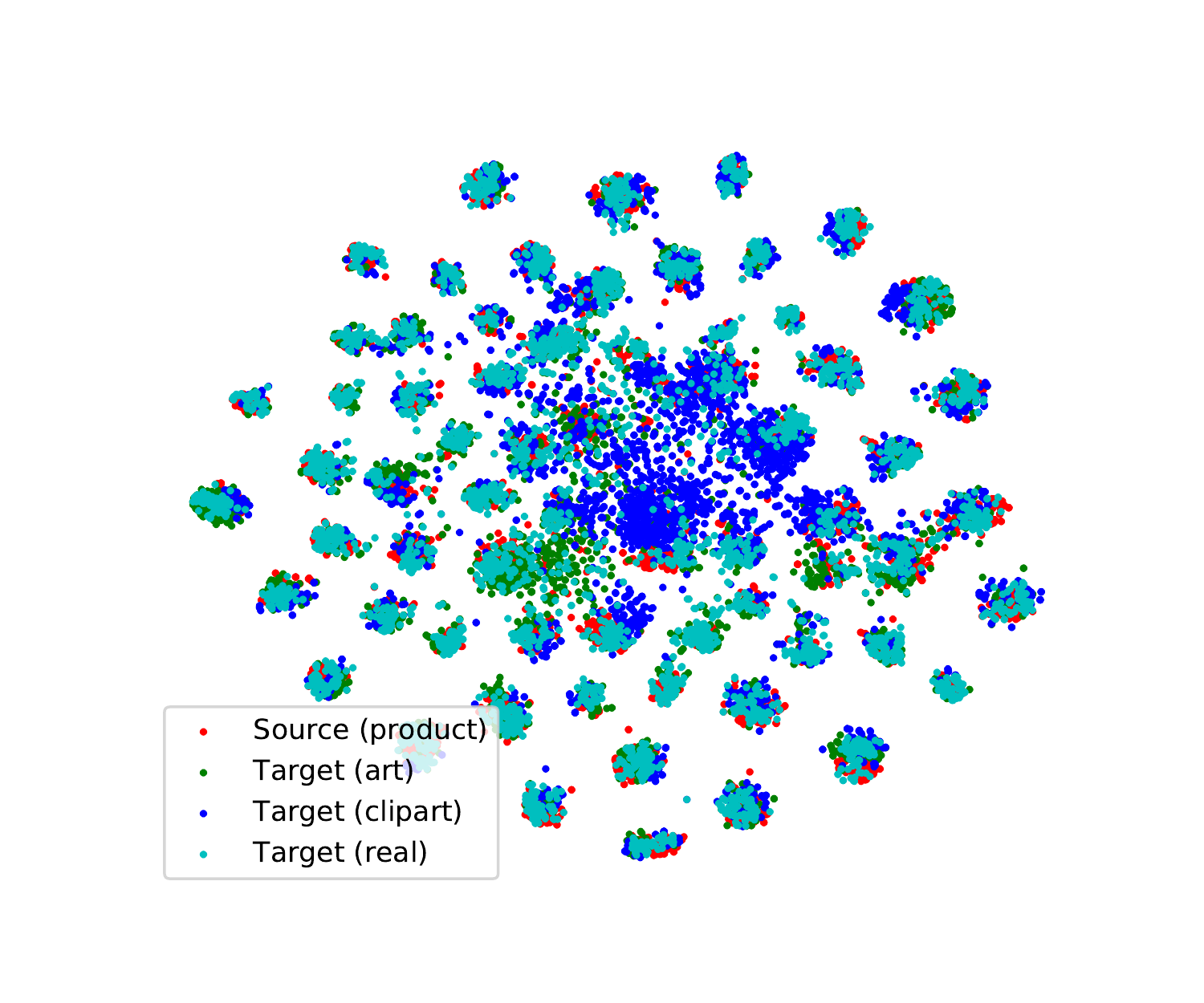} & \includegraphics[width=.5\linewidth]{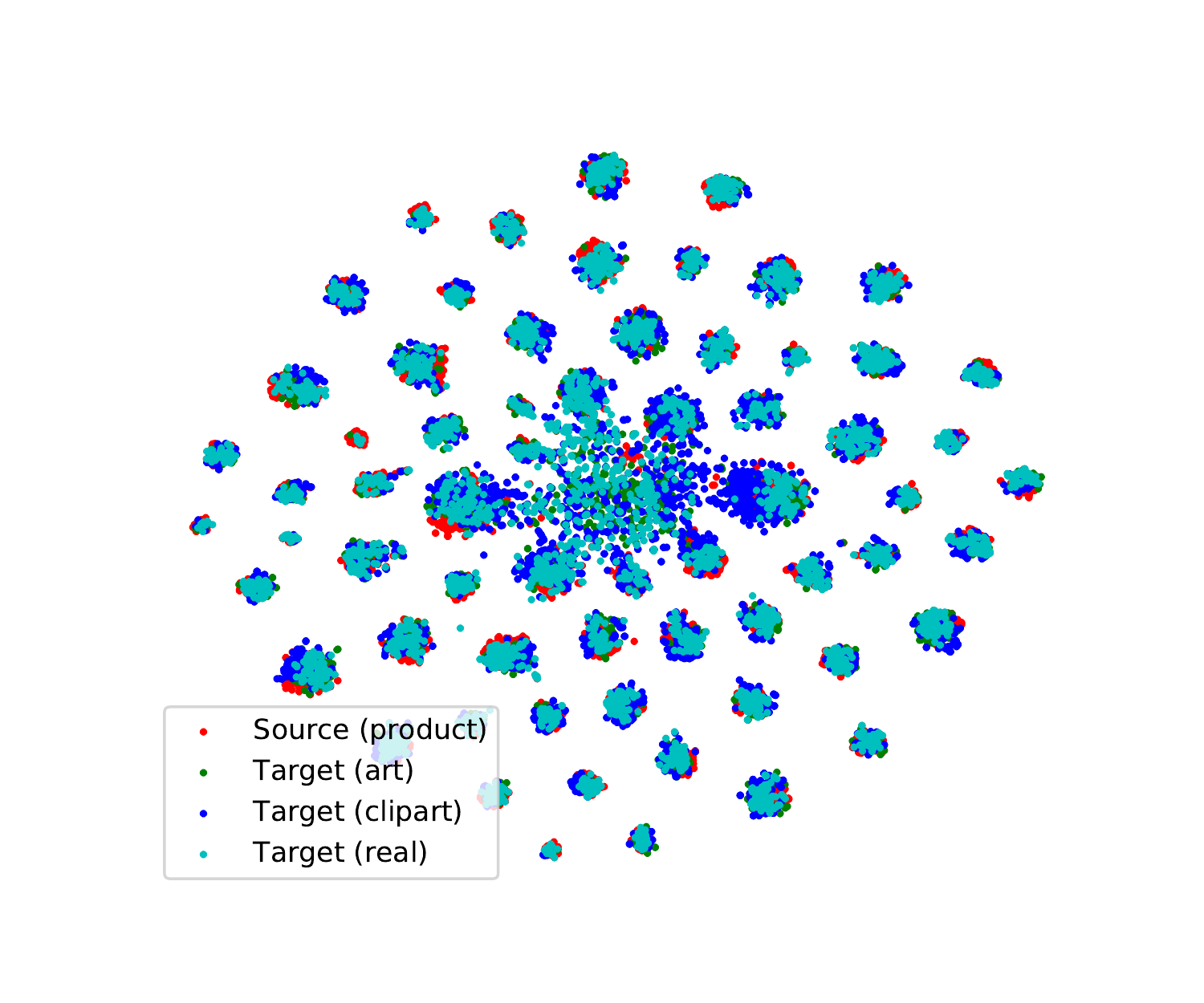} \\
    \end{tabular}
    \vspace{-.12in}
    \captionof{figure}{\textit{t}-SNE plots of the feature embeddings with Product $\rightarrow$ \textit{rest} in Office-Home. Left: \ours. Right: \domours.}
    \label{tab:tsne}
    \vspace{-5mm}
\end{table}

\begin{figure}[!h]
    \centering
    \includegraphics[width=\linewidth]{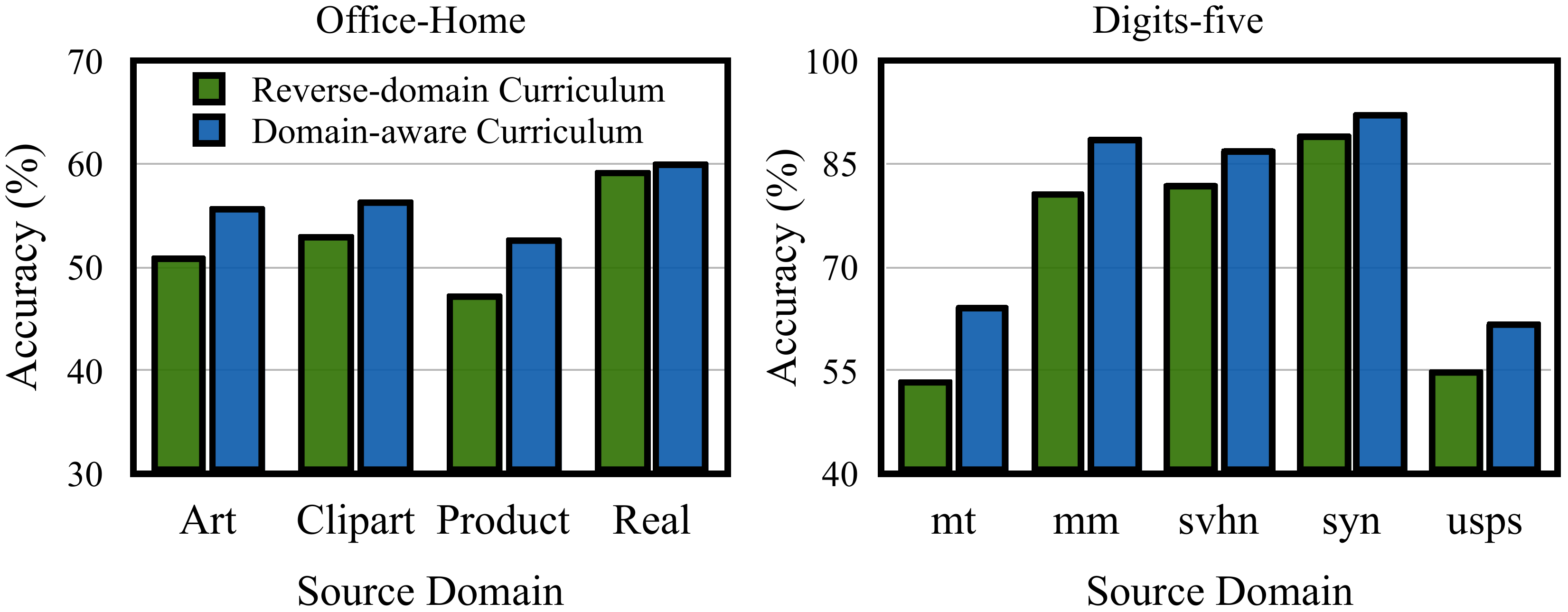}
    \vspace{-.22in}
    \caption{Comparison of the DCL with the \textit{reverse}-domain curriculum model on Office-Home and Digits-Five. In the reverse-domain curriculum model the order of selection of target domains is exactly opposite to that of the DCL model.}
    \label{fig:reverse_curriculum}
    \vspace{-5mm}
\end{figure}

To demonstrate that the order of target domains selection in the DCL indeed makes a difference, we consider a reverse-domain curriculum learning where the hardest domain is selected first, followed by the less hard ones. To this end, we train two models: i) Baseline\textsuperscript{$\dagger$}+DCL; and ii) Baseline\textsuperscript{$\dagger$}+Rev-DCL and compare their performances in the Fig.~\ref{fig:reverse_curriculum}. In both the datasets we observe the same phenomenon that the reverse-curriculum being detrimental. This once again re-establishes the importance of the proposed DCL in the MTDA setting.

\vspace{-1mm}
\subsection{Comparison with State-of-The-Art}
\vspace{-1mm}
\label{sec:sota}
We compare our proposed method and its variants with several state-of-the-art methods that are designed exclusively for the MTDA as well as the STDA methods that can be extended and used in the MTDA setting. In the main paper we only report the results for the Office-31, Office-Home and DomainNet experiments. Due to lack of space we report the numbers for Digits-five and PACS in the Supp. Mat.

In Tab.~\ref{tab:SOTA-Office31-Home} we report the numbers for Office-31 and Office-Home for single-target, target-combined and multi-target setting. The single-target setting denotes training single-source to single-target models, the target-combined means treating all the target domains as one aggregated target, while the multi-target setting comprise of training a single model for single-source to multiple-targets. As can be observed, in all the settings our proposed \ours and \domours outperform all the state-of-the-art methods. Specifically, for the Office-31, our \ours without using domain labels is already 2.4\% better than the HGAN~\cite{yang2020heterogeneous}, which is a MTDA method exploiting domain labels for feature aggregation with a single GCN classifier besides pseudo-labeling. This highlights the importance of having a co-teaching strategy with two classifiers and curriculum learning for counteracting the impact of noisy pseudo-labels in the GCN framework. We also observed that incorporating domain information following the proposed DCL strategy improves the performance in the Office-Home, with the \domours achieving 5.5\% improvement over MT-MTDA~\cite{nguyen2020unsupervised}, a MTDA method that also utilizes domain labels. Finally, as can be seen from the Tab.~\ref{tab:SOTA-DomainNet}, the \domours advances the state-of-the-art results for the very challenging DomainNet dataset by a non-trivial margin of 5.6\%. This further verifies the effectiveness of our proposed methods for addressing the MTDA.

\textbf{Overcoming negative transfer.} Careful inspection of the Tab.~\ref{tab:SOTA-Office31-Home} tells us that the single-target DA methods always outperform the same STDA method when applied in the multi-target setting. For e.g., CDAN is 4.3\% better in the single-target than in the multi-target setting. The drop in performance for the multi-target setting clearly hints at the fact that \textit{negative transfer}~\cite{peng2019domain, chen2019blending} is quite prevalent in the MTDA, despite having access to more data. Contrarily, our proposed \ours when applied to both the settings fares equally well for the Office-Home and outperforms the single-target counterpart by 0.6\% for the Office-31. This once again shows that the design choices made in our \ours and \domours lead to learning more robust domain-invariant features and provide resilience against negative transfer.

%% file: supp.tex
In this supplementary material, we provide more implementation details, discussion of the proposed \ours, and additional experimental results. In details, we provide the pseudo-code  algorithms of the proposed \ours and \domours in Sec.~\ref{sec:algo}. We highlight the key differences between \ours and prior works in Sec.~\ref{sec:discussion}. The details of datasets and implementation are provided in Sec.~\ref{sec:dataset} and Sec.~\ref{sec:imple-details}, respectively. The additional experimental results are reported in Sec.~\ref{sec:add_exp}.

\section{Algorithms}
\label{sec:algo}
In this section we provide the pseudo-code algorithms for the proposed \ours (see Sec.~3.2 of the main paper) and \domours (see Sec.~3.3 of the main paper) in the Alg.~\ref{algo:cgct} and Alg.~\ref{algo:dcgct}, respectively. Note that the \textit{adaptation stage} in the Alg.~\ref{algo:dcgct} can be replaced by any desired single-target domain adaptation (STDA) method of choice, thereby, making the proposed DCL flexible to a wide variety of STDA methods.

\begin{algorithm}
\small
\SetAlgoLined
\LinesNumbered
\SetCommentSty{mycommfont}
\SetKwInOut{Require}{require}
\SetKwInOut{Output}{output}
\SetKwComment{Comment}{$\triangleright$\ }{}
\Require{number of target domains $N$, classes $n_c$}
\Require{source dataset $\mathcal{S}$; combined target dataset $\mathcal{T}$}
\Require{hyper-parameters $B$, $\tau$, $K$, $K', \lambda_{edge}, \lambda_{node}, \lambda_{adv}$}
\Require{networks $F$, $D$, $G_{mlp}$, $f_{edge}$, $f_{node}$ with parameters $\theta, \psi, \phi, \varphi, \varphi'$, respectively. The $f_{edge}$ and $f_{node}$ form the $G_{gcn}$.}

\nonl\underline{\textbf{Step 1}: \textit{Pre-training on the source dataset}}\\
 \While{$\ell_{ce}$ has not converged}{
    ($\mathbf{x}_{s,i}, y_{s,i})^{B}_{i=1} \sim \mathcal{S}$\\
    update $\theta$, $\phi$ by $\Min_{\theta, \phi} \ell^{mlp}_{ce}$\\
 }
 \nonl\underline{\textbf{Step 2}: \textit{Curriculum learning}}\\
 $\hat{\mathcal{S}}^{0} \leftarrow \mathcal{S}$ \\
 \DontPrintSemicolon $Q \leftarrow N$ \Comment*{Total \# curriculum steps}
 \For{$q$ $\text{in}$ (0 : $Q - 1$)}{
    \DontPrintSemicolon \Comment*{Curriculum step}
    \nonl\underline{\textbf{Stage 1}: \textit{Adaptation stage}}\\
    \For{$k \; \text{in} \; (1 : K)$}{
        $\hat{\mathcal{B}}^{q}_{s} \leftarrow (\mathbf{x}_{s,i}, y_{s,i})^{B}_{i=1} \sim \hat{\mathcal{S}}^{q}$ \\
        $\hat{\mathcal{B}}^{q}_{t} \leftarrow (\mathbf{x}_{t,i})^{B}_{i=1} \sim \mathcal{T}$\\
        $\hat{y} \leftarrow \texttt{softmax} (G_{mlp}(F(\mathbf{x})))$ \\
        $\bar{y} \leftarrow \texttt{softmax} (G_{gcn}(F(\mathbf{x})))$\\
        $\hat{d} \leftarrow \texttt{sigmoid} (D(F(\mathbf{x})))$\\
        update $\psi$ by $\Min_{\psi} \lambda_{adv} \ell_{adv}$\\
        update $\theta$, $\phi$ by $\Min_{\theta, \phi} \ell^{mlp}_{ce} - \lambda_{adv} \ell_{adv}$\\
        update $\theta$, $\varphi, \varphi'$ by $\Min_{\theta, \varphi, \varphi'} \lambda_{edge} \ell^{edge}_{bce} + \lambda_{node} \ell^{node}_{ce}$\\
    }
    \nonl\underline{\textbf{Stage 2}: \textit{Pseudo-labeling stage}}\\
    $\mathcal{D}^{q}_{t} \leftarrow \{\}$ \Comment*{Empty list}
    \For{$\mathbf{x}_{t,j} \in \mathcal{T}$}{
        $w_j \leftarrow \Max_{c \in n_c} p(\bar{y}_{t,j} = c | \mathbf{x}_{t,j})$\\
        \If{$w_j > \tau$}{
            $\mathcal{D}^{q}_{t} \leftarrow \mathcal{D}^{q}_{t} || \{(\mathbf{x}_{t,j}, \ArgMax_{c \in n_c}p(\bar{y}_{t,j} = c | \mathbf{x}_{t,j})) \}$ \Comment*{Append}
        }
    }
    $\hat{\mathcal{S}}^{q+1} \leftarrow \mathcal{S} \cup \mathcal{D}^{q}_{t}$ \Comment*{Pseudo-source}
 }
 \nonl\underline{\textbf{Step 3}: \textit{Fine-tuning on pseudo-source dataset}}\\
 \For{$k' \; \text{in} \; (1 : K')$}{
    $(\mathbf{x}_{s,i}, y_{s,i})^{B}_{i=1} \sim \hat{\mathcal{S}}^{Q}$\\
    update $\theta$, $\phi$ by $\Min_{\theta, \phi} \ell^{mlp}_{ce}$\\
 }
 \caption{Training Procedure of Curriculum Graph Co-Teaching (CGCT)}
 \label{algo:cgct}
\end{algorithm}

\begin{algorithm}
\small
\SetAlgoLined
\LinesNumbered
\SetCommentSty{mycommfont}
\SetKwInOut{Require}{require}
\SetKwInOut{Output}{output}
\SetKwComment{Comment}{$\triangleright$\ }{}
\Require{number of target domains $N$, classes $n_c$}
\Require{source dataset $\mathcal{S}$; target dataset $\mathcal{T} = \{\mathcal{T}_j\}^{N}_{j=1}$}
\Require{hyper-parameters $B$, $\tau$, $K$, $K', \lambda_{edge}, \lambda_{node}, \lambda_{adv}$}
\Require{networks $F$, $D$, $G_{mlp}$, $f_{edge}$, $f_{node}$ with parameters $\theta, \psi, \phi, \varphi, \varphi'$, respectively. The $f_{edge}$ and $f_{node}$ form the $G_{gcn}$.}

\nonl\underline{\textbf{Step 1}: \textit{Pre-training on the source dataset}}\\
 \While{$\ell_{ce}$ has not converged}{
    ($\mathbf{x}_{s,i}, y_{s,i})^{B}_{i=1} \sim \mathcal{S}$\\
    update $\theta$, $\phi$ by $\Min_{\theta, \phi} \ell^{mlp}_{ce}$\\
 }
 \nonl\underline{\textbf{Step 2}: \textit{Curriculum learning}}\\
 $\hat{\mathcal{S}}^{0} \leftarrow \mathcal{S}$ and $\hat{\mathcal{T}}^{0} \leftarrow \{\mathcal{T}_j\}^{N}_{j=1}$ \\
 \DontPrintSemicolon $Q \leftarrow N$ \Comment*{Total \# curriculum steps}
 \For{$q$ $\text{in}$ (0 : $Q - 1$)}{
    \DontPrintSemicolon \Comment*{Curriculum step}
    \DontPrintSemicolon $\mathcal{H} \leftarrow \{\}$ \Comment*{Empty list}
    \nonl\underline{\textbf{Stage 1}: \textit{Domain selection stage}}\\
    \For{$\mathcal{T}_j \; \text{in} \; \hat{\mathcal{T}}^{q}$}{
        compute $H(\mathcal{T}_j)$ as in Eqn. 12\\
        $\mathcal{H} \leftarrow \mathcal{H} \; || \; H(\mathcal{T}_j)$ \Comment*{Append}
    }
    $\mathbb{D}^{q} \leftarrow \Argmin_{j} \mathcal{H}$ \Comment*{Chosen domain}
    \nonl\underline{\textbf{Stage 2}: \textit{Adaptation stage}}\\
    \For{$k \; \text{in} \; (1 : K)$}{
        $\hat{\mathcal{B}}^{q}_{s} \leftarrow (\mathbf{x}_{s,i}, y_{s,i})^{B}_{i=1} \sim \hat{\mathcal{S}}^{q}$ \\
        $\hat{\mathcal{B}}^{q}_{t} \leftarrow (\mathbf{x}_{t,i})^{B}_{i=1} \sim \mathcal{T}_{\mathbb{D}^{q}}$\\
        $\hat{y} \leftarrow \texttt{softmax} (G_{mlp}(F(\mathbf{x})))$ \\
        $\bar{y} \leftarrow \texttt{softmax} (G_{gcn}(F(\mathbf{x})))$\\
        $\hat{d} \leftarrow \texttt{sigmoid} (D(F(\mathbf{x})))$\\
        update $\psi$ by $\Min_{\psi} \lambda_{adv} \ell_{adv}$\\
        update $\theta$, $\phi$ by $\Min_{\theta, \phi} \ell^{mlp}_{ce} - \lambda_{adv} \ell_{adv}$\\
        update $\theta$, $\varphi, \varphi'$ by $\Min_{\theta, \varphi, \varphi'} \lambda_{edge} \ell^{edge}_{bce} + \lambda_{node} \ell^{node}_{ce}$\\
    }
    \nonl\underline{\textbf{Stage 3}: \textit{Pseudo-labeling stage}}\\
    $\mathcal{D}^{\mathbb{D}^{q}}_{t} \leftarrow \{\}$ \Comment*{Empty list}
    \For{$\mathbf{x}_{t,j} \in \mathcal{T}_{\mathbb{D}^{q}}$}{
        $w_j \leftarrow \Max_{c \in n_c} p(\bar{y}_{t,j} = c | \mathbf{x}_{t,j})$\\
        \If{$w_j > \tau$}{
            $\mathcal{D}^{\mathbb{D}^{q}}_{t} \leftarrow \mathcal{D}^{\mathbb{D}^{q}}_{t} || \{(\mathbf{x}_{t,j}, \ArgMax_{c \in n_c}p(\bar{y}_{t,j} = c | \mathbf{x}_{t,j})) \}$ \Comment*{Append}
        }
    }
    $\hat{\mathcal{S}}^{q+1} \leftarrow \hat{\mathcal{S}}^{q} \cup \mathcal{D}^{\mathbb{D}^{q}}_{t}$ \Comment*{Pseudo-source}
    $\hat{\mathcal{T}}^{q+1} = \hat{\mathcal{T}}^{q} \; \text{\textbackslash } \; \mathcal{T}_{\mathbb{D}^{q}}$\\
 }
 \nonl\underline{\textbf{Step 3}: \textit{Fine-tuning on pseudo-source dataset}}\\
 \For{$k' \; \text{in} \; (1 : K')$}{
    $(\mathbf{x}_{s,i}, y_{s,i})^{B}_{i=1} \sim \hat{\mathcal{S}}^{Q}$\\
    update $\theta$, $\phi$ by $\Min_{\theta, \phi} \ell^{mlp}_{ce}$\\
 }
 \caption{Training Procedure of Domain-aware Curriculum Graph Co-Teaching (D-CGCT)}
 \label{algo:dcgct}
\end{algorithm}

\section{Discussion}
\label{sec:discussion}
Here we highlight the keys differences between the \ours and PGL~\cite{luo2020progressive} as well as the dual classifier-based methods~\cite{han2018co, saito2018maximum}. The PGL~\cite{luo2020progressive} exploits the graph learning framework in an episodic fashion to obtain pseudo-labels for the unlabeled target samples, which are then used to bootstrap the model by training on the pseudo-labeled target data. While our proposed method is similar in spirit to the episodic training in~\cite{luo2020progressive}, we do not solely rely on the GCN to obtain the pseudo-labels. We conjecture that due to the fully-connected nature of the graph and lack of target labels, the GCN will be prone to accumulate features of dissimilar neighbours, thereby, resulting in the erroneous label propagation. To address this peculiarity, we propose to resort to the co-teaching paradigm, where the $G_{mlp}$ is exploited to train the $f_{edge}$ network. As the two classifiers will capture different aspects of training~\cite{han2018co}, it will prevent the $f_{edge}$ to be trained with the same erroneous pseudo-labels as the $f_{node}$. We validate this conjecture empirically, where a network with a single GCN classifier with pseudo-labels performs sub-optimally compared to \ours (see Tab.~5 row 7 of the main paper). Finally, the dual classifier-based methods maintain two classifiers to identify and filter either harder target samples~\cite{saito2018maximum} or noisy samples~\cite{han2018co}. Contrarily, we maintain $G_{mlp}$ and $G_{gcn}$ to provide feedback to each other by exploiting the key observation that each classifier learns different patterns during training. Furthermore, given the intrinsic design of the $G_{gcn}$, we also do away with an extra adhoc loss of keeping the weights of two networks different.

\section{Datasets}
\label{sec:dataset}
\textit{Digits-five}~\cite{xu2018deep} is composed of five domains that are drawn from the: i) grayscale handwritten digits MNIST~\cite{lecun1998gradient} (\textbf{mt}); ii) a coloured version of \textbf{mt}, called as MNIST-M~\cite{ganin2016domain} (\textbf{mm}); iii) USPS~\cite{friedman2001elements} (\textbf{up}), which is a lower resolution, 16$\times$16, of the handwritten digits \textbf{mt}; iv) a real-world dataset of digits called SVHN~\cite{netzer2011reading} (\textbf{sv}); and v) a synthetically generated dataset \textit{Synthetic Digits}~\cite{ganin2016domain} (\textbf{sy}). Following the protocol of~\cite{chen2019blending}, we sub-sample 25,000 and 9,000 samples from the training and test sets of \textbf{mt}, \textbf{mm}, \textbf{sv} and \textbf{sy} and use as train and test sets, respectively. For the \textbf{up} domain we use all the 7,348 training and 1,860 and test samples, for our experiments. All the images are re-scaled to a 28$\times$28 resolution.

\textit{Office31}~\cite{saenko2010adapting} is a standard visual DA dataset comprised of three domains: Amazon, DSLR and Webcam. The dataset consists of 31 distinct object categories with a total of 4,652 samples.

\textit{Office-Home}~\cite{venkateswara2017Deep} is a relatively newer DA benchmark that is larger than Office31 and is composed of four different visual domains: Art, Clipart, Product and Real. It consists of 65 object categories and has 15,500 images in total.

\textit{PACS}~\cite{li2017deeper} is another visual DA benchmark that also consists of four domains: Photo (P), Art Painting (A), Cartoon (C) and Sketch (S). This dataset is captured from 7 object categories and has 9,991 images in total.

\textit{DomainNet}~\cite{peng2019moment} is the most challenging and very large scale DA benchmark, which has six different domains: Clipart (C), Infograph (I), Painting (P), Quickdraw (Q), Real (R) and Sketch (S). It has around \textbf{0.6 million} images, including both train and test images, and has 345 different object categories. We use the official training and testing splits, as mentioned in~\cite{peng2019domain}, for our experiments.

\begin{table*}[!t]
    \centering
    \setlength{\tabcolsep}{1.2pt}
    \begin{tabular}{c c}
        \includegraphics[width=.45\linewidth]{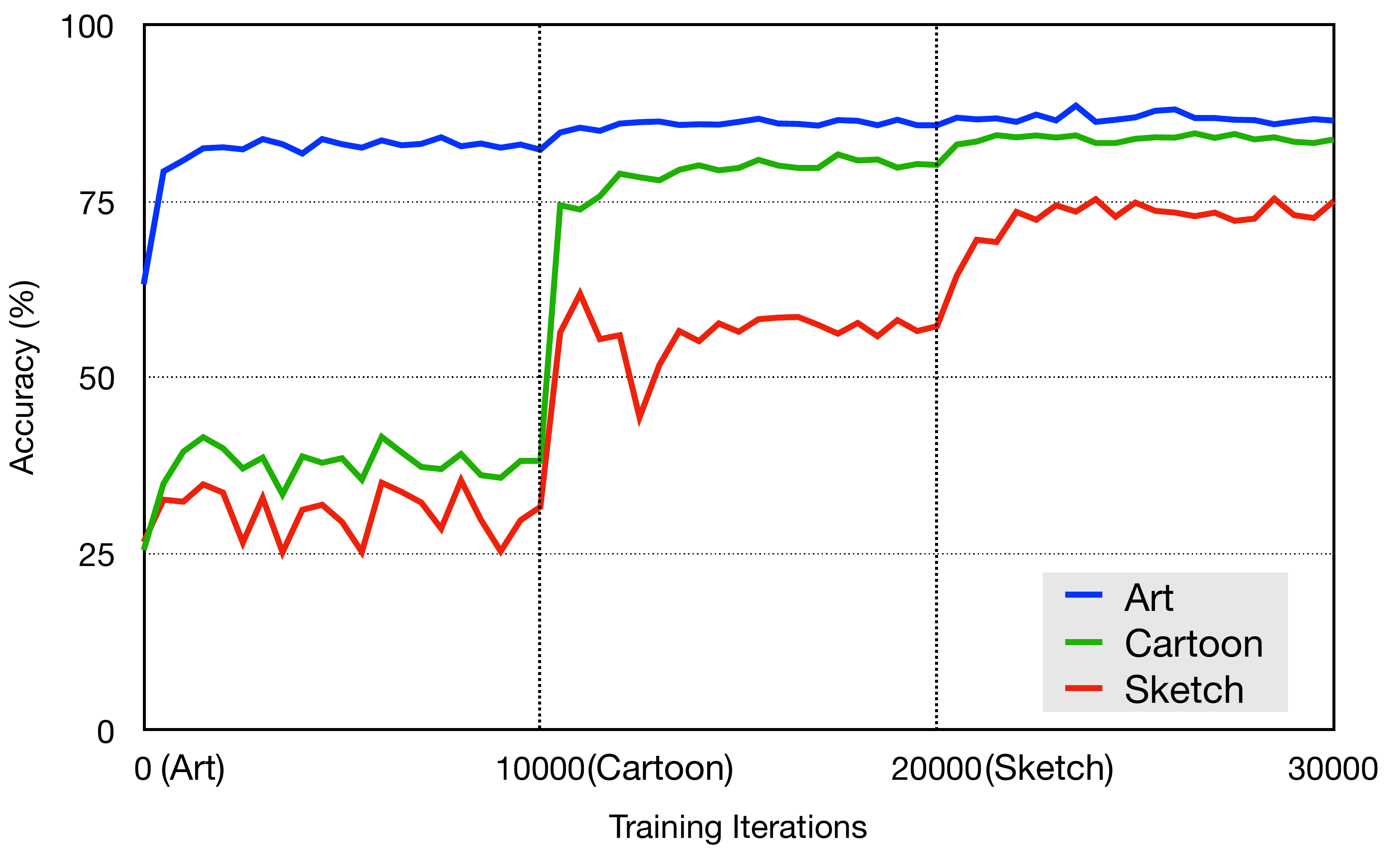} &
        \includegraphics[width=.45\linewidth]{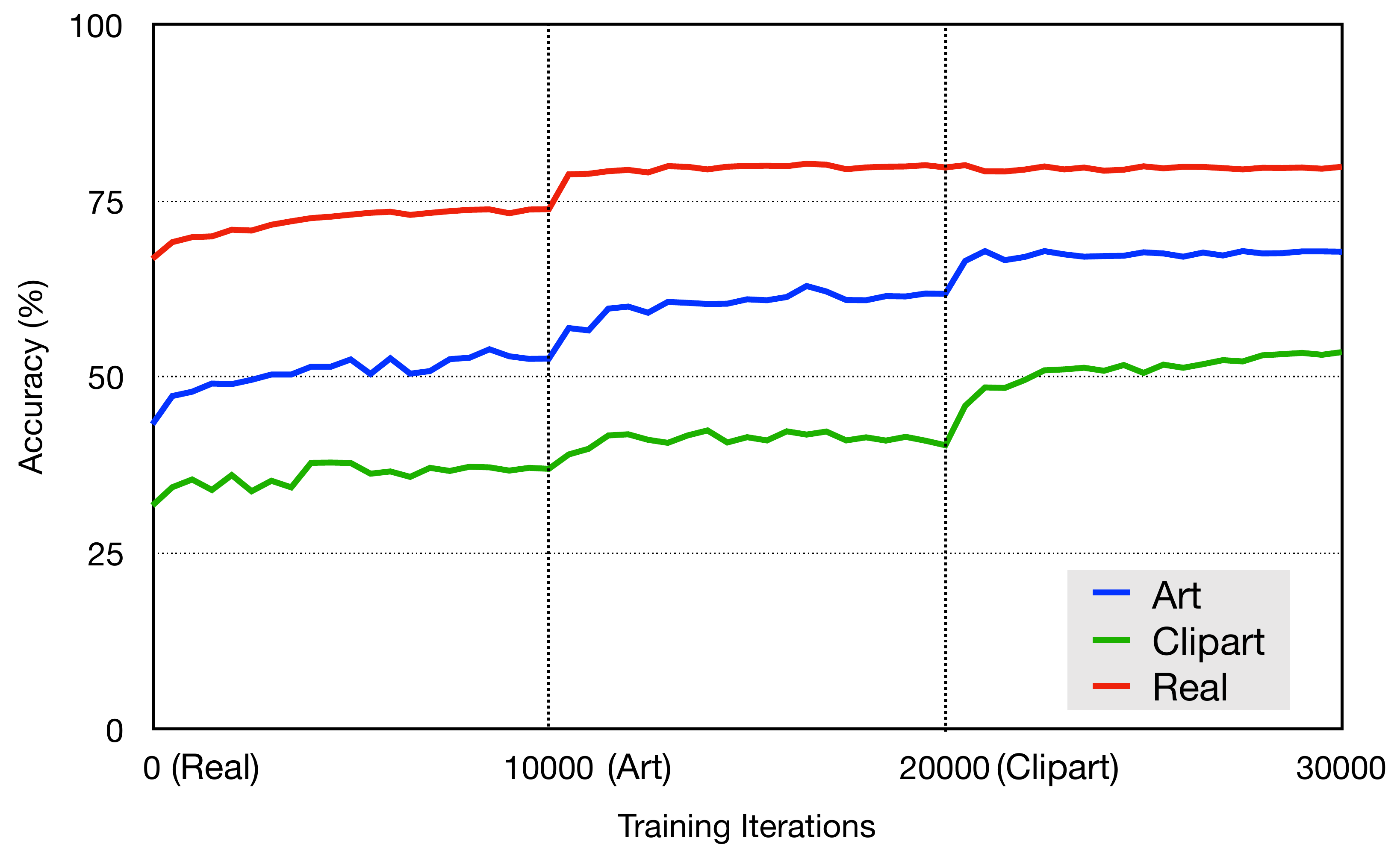}\\
        (a) Photo $\rightarrow$ \textit{rest} in the PACS & (a) Product $\rightarrow$ \textit{rest} in the Office-Home \\
    \end{tabular}
    \vspace{-.12in}
    \captionof{figure}{The classification accuracy line plots with the \domours using ResNet-50 as the backbone. At each indicated training iteration in the x-axis, a new target domain (shown in brackets) is selected for adaptation.}
    \label{fig:lines}
\end{table*}

\begin{table}[!h]
    \centering
    \small
    \setlength{\tabcolsep}{3.8pt}
    \begin{tabular}{lcccc}
         \hline
         Layer & \thead{\makecell{$k_{size}$, $C_{in}$, $C_{out}$,\\ $st$, $pad$}} & \thead{\makecell{IN/BN}} & \thead{\makecell{Non-\\linearity}} & \thead{\makecell{Dropout}} \\
         \hline
        \makecell[l]{\textbf{Feature-} \\\textbf{extractor}} & \\
        Conv1 &  (5, 3, 32, 1, 0) & IN/BN & ReLU & 0.2\\
        Maxpool2d & (2, -, -, 2, -) & - & -& -\\
        Conv2 &  (5, 32, 64, 1, 0) & BN & ReLU & 0.2\\
        Maxpool2d & (2, -, -, 2, -) & - & -& -\\
        FC3 & (-, 64*4*4, 100, -, -) & BN & ReLU & 0.2\\
        FC4 & (-, 100, 100, -. -) & BN & ReLU & -\\
        \hline
        \makecell[l]{\textbf{Classifier}} & \\
        FC\_out & (-, 100, 10, -, -) & - & - & -\\
        \hline
        \makecell[l]{\textbf{Domain-}\\\textbf{Discriminator}} & \\
        D\_FC1 & (-, 100*10, 100, -, -) & - & ReLU & 0.5\\
        D\_FC2 & (-, 100, 100, -, -) & - & ReLU & 0.5\\
        D\_FC3 & (-, 100, 1, -, -) & - & - & -\\
         \hline
    \end{tabular}
    \vspace{-.12in}
    \caption{The network architecture for the baseline~\cite{long2018conditional} used in the Digits-five experiments. Kernel size ($k_{size}$); in channels ($C_{in}$); out channels ($C_{out}$); stride ($st$); and padding ($pad$). IN stands for instance normalization. The input image resolution is 28 $\times$ 28 $\times$ 3.}
    \label{tab:digits_arch}
    \vspace{-3mm}
\end{table}

\begin{table*}[!h]
    \centering
    \small
    \def\arraystretch{1.1}
    \setlength{\tabcolsep}{4.0pt}
    \begin{tabular}{l|lcccccc}
    \specialrule{1.5pt}{1pt}{1pt}
    & & \multicolumn{6}{c}{Digits-five} \\
    \cline{3-8}
    \thead{Setting} & \thead{Model} & \thead{mt $\rightarrow$ mm,sv,sy,up} & \thead{mm $\rightarrow$ mt,sv,sy,up} & \thead{sv $\rightarrow$ mm,mt,sy,up} & \thead{sy $\rightarrow$ mm,sv,mt,up} & \thead{up $\rightarrow$ mm,sv,sy,mt} & \textbf{Avg} (\%)\\
    \specialrule{1.5pt}{1pt}{1pt}
    \multirow{8}{*}{\makecell{Target \\Combined}} & Source only & 26.9 & 56.0 & 67.2 & 73.8 & 36.9 & 52.2 \\
    & ADDA~\cite{Hoffman:Adda:CVPR17} & 43.7 & 55.9 & 40.4 & 66.1 & 34.8 & 48.2 \\
    & DAN~\cite{long2015learning} & 31.3 & 53.1 & 48.7 & 63.3 & 27.0 & 44.7 \\
    & GTA~\cite{sankaranarayanan2018generate} & 44.6 & 54.5 & 60.3 & 74.5 & 41.3 & 55.0 \\
    & RevGrad~\cite{ganin2016domain} & 52.4 & 64.0 & 65.3 & 66.6 & 44.3 & 58.5 \\
    & AMEAN~\cite{chen2019blending} & \textbf{56.2} & 65.2 & 67.3 & 71.3 & 47.5 & 61.5 \\
    & CDAN~\cite{long2018conditional} & 53.0 & 76.3 & 65.6 & 81.5 & \textbf{56.2} & 66.5\\
    & \textbf{\ours} & 54.3 & \textbf{85.5} & \textbf{83.8} & \textbf{87.8} & 52.4 & \textbf{72.8} \\
    \hline
    \multirow{3}{*}{\makecell{Multi-\\Target}} & CDAN~\cite{long2018conditional} & 53.7 & 76.2 & 64.4 & 80.3 & 46.2 & 64.2 \\
    & \textbf{CDAN} + \textbf{DCL} & 62.0 & 87.8 & 87.8 & 92.3 & \textbf{63.2} & 78.6\\
    & \textbf{\domours} & \textbf{65.7} & \textbf{89.0} & \textbf{88.9} & \textbf{93.2} & 62.9 & \textbf{79.9} \\
    \specialrule{1.5pt}{1pt}{1pt}
    \end{tabular}
    \vspace{-.12in}
    \caption{Comparison with the state-of-the-art methods on the Digits-five. ``Target Combined'' indicates methods are performed on one source to one combined target domain. ``Multi-Target'' indicates methods are performed on one source to multi-target setting. Our proposed models are highlighted in bold.}
    \label{tab:SOTA-digits}
\end{table*}

\section{Implementation Details}
\label{sec:imple-details}
\textbf{General Setting.} To be fairly comparable with the state-of-the-art methods, we adopted comparable backbone feature extractors in the corresponding experiments and datasets. For Digits-five, we have used a small convolutional network as the backbone feature extractor (see Tab.~\ref{tab:digits_arch}), which is adapted from~\cite{chen2019blending} and includes two \textit{conv} layers and two \textit{fc} layers. We trained the model using a Stochastic Gradient Descent (SGD) optimizer with an initial learning rate of 1e-3. For the rest of the datasets, we have adoptd ResNet~\cite{he2016deep} based feature extractors. Specifically, for the ablation studies on Office-Home, we have used ResNet-18 as the backbone network. For the state-of-the-art comparisons on Office31, PACS and Office-Home we have used ResNet-50. For the DomainNet, we have utilized ResNet-101 as used by the competitor methods.
Similarly to the Digits-five, SGD optimizer is used with an initial learning rate of 1e-3 and is decayed exponentially. Each curriculum step consists of $K = 10,000$ training iterations for all the datasets, except the DomainNet, where $K=50,000$ due to large size of the dataset. The final fine-tuning step is trained with $K'=15,000$ iterations for all datasets.

\textbf{GCN architecture.} We have implemented $f_{node}$ network with 2 conv layers followed by a Batch Normalization (BN) layer and ReLU activation, except the final layer. The first layer takes as input image features concatenated with the context of the mini-batch, \textit{i.e.}, the aggregated features of other images in a mini-batch (based on the affinity matrix estimated by the $f_{edge}$). The second conv layer outputs the logits that are equal to the number of classes $n_c$. We have used 1x1 convolution kernels in the $f_{node}$. Similarly, we have implemented the $f_{edge}$ network with 3 conv layers and 1x1 kernels, where the first two layers are followed by the BN layers and ReLU activations, except the last. The third conv layer has a single channel as output, thus, representing the similarity scores between samples in a mini-batch.

\begin{table*}[!h]
    \centering
    \small
    \def\arraystretch{1.1}
    \setlength{\tabcolsep}{7.0pt}
    \begin{tabular}{l|lccccccc}
    \specialrule{1.5pt}{1pt}{1pt}
    & & \multicolumn{7}{c}{PACS} \\
    \cline{3-9}
    Setting&Model & A $\rightarrow$ S & A $\rightarrow$ C & A $\rightarrow$ P & P $\rightarrow$ S & P $\rightarrow$ C & P $\rightarrow$ A & \textbf{Avg} (\%)\\
    \specialrule{1.5pt}{1pt}{1pt}
    \multirow{4}{*}{Target Combined} & MSTN~\cite{zhang2015deep} & 70.4 & 71.2 & 96.2 & \textbf{55.9} & \textbf{49.1} & 70.8 &  68.9\\
    & ADDA~\cite{Hoffman:Adda:CVPR17} & 65.3 & 68.0 & 96.0 & 48.8 & 47.1 & 67.3 & 65.4 \\
    & CDAN~\cite{long2018conditional} & 56.8 & 61.1 & 95.9 & 55.7 & 53.8 & 49.4 & 62.1 \\
     & \textbf{\ours} & \textbf{70.5} & \textbf{75.4} & \textbf{98.3} & 44.6 & 44.3 & \textbf{81.7} & \textbf{69.1} \\
    \cline{1-9}
    \multirow{4}{*}{Multi-Target} & CDAN~\cite{long2018conditional} & 75.9 & 81.9 & 95.4 & 51.3 & 61.7 & 65.0 & 71.9 \\
    & HGAN~\cite{yang2020heterogeneous} &  72.1 & 78.3 & 97.7 & 70.8 & 62.8 &  78.8 & 76.8 \\
    & \textbf{CDAN} + \textbf{DCL} & 68.7 & 89.0 & 98.8 & 61.2 & \textbf{82.9} & \textbf{89.8} & 81.7\\
    & \textbf{\domours} & \textbf{84.6} & \textbf{90.2} & \textbf{99.4} & \textbf{76.5} & 82.4 & 88.6 & \textbf{87.0}\\
    \specialrule{1.5pt}{1pt}{1pt}
    \end{tabular}
    \vspace{-.12in}
    \caption{Comparison with the state-of-the-art methods on the PACS. All methods use the ResNet-50 as the backbone.
    ``Target Combined'' indicates methods are performed on one source to one combined target domain. ``Multi-Target'' indicates methods are performed on one source to multi-target setting. Our proposed models are highlighted in bold.}
    \label{tab:SOTA-PACS}
\end{table*}

\section{Additional Experiments}
\label{sec:add_exp}

\subsection{Ablations}
To explain why the step-by-step adaptation in the proposed DCL better addresses the alleviation of the larger domain-shifts in the MTDA setting, we plot the classification accuracy with the \domours in Fig.~\ref{fig:lines}. As can be observed from the Fig.~\ref{fig:lines} (a), for Photo $\rightarrow$ \textit{rest} setting in the PACS, when the adaptation first begins with the Art as target, the performance of the model on the \textit{unseen} Cartoon domain simultaneously improves in the first 10k iterations (or the $1^{\text{st}}$ curriculum step), despite the network not seeing any sample from the Cartoon domain. This phenomenon is even vividly noticeable in the second curriculum step, where the performance on the unseen Sketch largely increases when the Cartoon is selected for adaptation. This in other words means that the domain-shift between the source (Photo) and the farthest target (Sketch) has already been considerably reduced by the time the Sketch enters the adaptation stage (from 20k iterations on wards). Thus, we empirically demonstrate the prime reason behind the DCL achieving superior performance over other state-of-the-art MTDA methods. Similar observations can also be noticed for the Office-Home. We depict the Product $\rightarrow$ \textit{rest} setting in the Fig.~\ref{fig:lines} (b).

\subsection{Comparison with the State-of-the-Art}

\begin{table*}[!t]
    \centering
    \small
    \def\arraystretch{1.1}
    \setlength{\tabcolsep}{7.0pt}
    \begin{tabular}{l|lccccccccc}
    \specialrule{1.5pt}{1pt}{1pt}
    & & \multicolumn{9}{c}{DomainNet} \\
    \cline{3-11}
    Setting&Model & R $\rightarrow$ S & R $\rightarrow$ C & R $\rightarrow$ I & R $\rightarrow$ P & P $\rightarrow$ S & P $\rightarrow$ R & P $\rightarrow$ C & P $\rightarrow$ I &  \textbf{Avg} (\%)\\
    \specialrule{1.5pt}{1pt}{1pt}
    \multirow{4}{*}{Target Combined} & MSTN~\cite{zhang2015deep} &  31.4 &  40.2 &  14.9 &  40.5 & 31.5 &  48.3 & 32.2 & 13.0 &  31.5 \\
    & ADDA~\cite{Hoffman:Adda:CVPR17} & 27.5 & 33.9 & 12.7 &  35.0 & 26.2 & 41.7 & 26.9 & 10.7 & 26.8 \\
    & CDAN~\cite{long2018conditional} & 40.8 & 52.7 & 21.5 & 48.7 & 37.8 & 57.8 & 44.1 & 17.7 & 40.1 \\
    & \textbf{\ours} &  \textbf{48.9} & \textbf{60.3} & \textbf{26.9} & \textbf{57.1} & \textbf{43.4} & \textbf{58.8} & \textbf{48.5} & \textbf{21.7} & \textbf{45.7} \\
    \cline{1-11}
    \multirow{4}{*}{Multi-Target} & CDAN~\cite{long2018conditional} & 40.7 & 51.9 & 22.5 & 49.0 & 39.6 & 57.9 & 44.6 & 18.4 & 40.6 \\
    & HGAN~\cite{yang2020heterogeneous} & 34.3 & 43.2 & 17.8 & 43.4 & 35.7 & 52.3 & 35.9 & 15.6 &  34.7 \\
    & \textbf{CDAN} + \textbf{DCL} & 45.2 & 58.0 & 23.7 & 54.0 & 45.0 & \textbf{61.5} & 50.7 & 20.3 & 44.8 \\
    & \textbf{\domours} &  \textbf{48.4} & \textbf{59.6} & \textbf{25.3} & \textbf{55.6} & \textbf{45.3} & 58.2 & \textbf{51.0} & \textbf{21.7} & \textbf{45.6} \\
    \specialrule{1.5pt}{1pt}{1pt}
    \end{tabular}
    \vspace{-.12in}
    \caption{Comparison with the state-of-the-art methods on the DomainNet. All methods use the ResNet-101 as the backbone.
    ``Target Combined'' indicates methods are performed on one source to one combined target domain. ``Multi-Target'' indicates methods are performed on one source to multi-target setting. Our proposed models are highlighted in bold.}
    \label{tab:SOTA-hgan-domainnet}
\end{table*}

\begin{table*}[!t]
    \centering
    \setlength{\tabcolsep}{.2pt}
    \begin{tabular}{c c c c}
        \includegraphics[width=.25\linewidth]{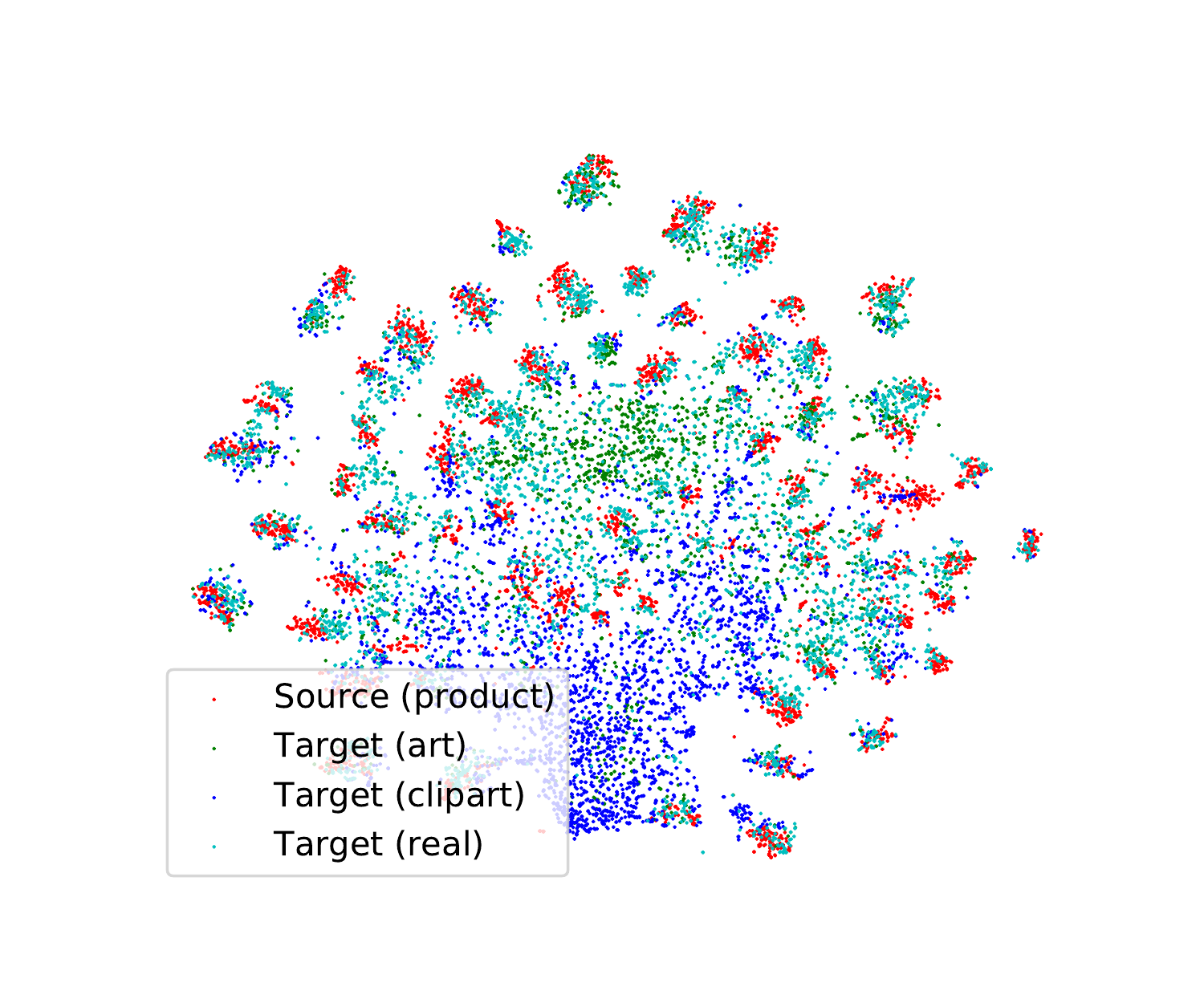} &
        \includegraphics[width=.25\linewidth]{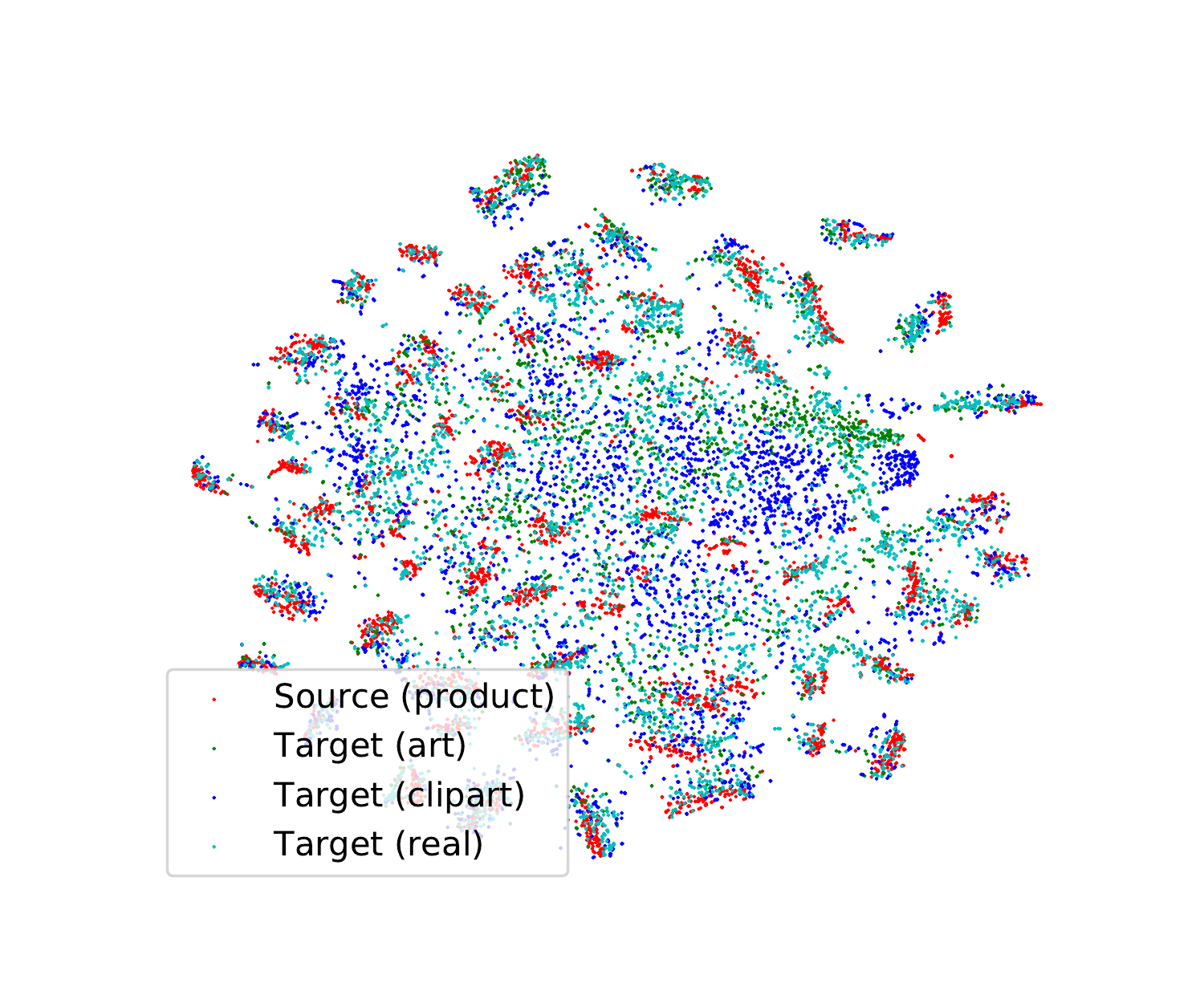} &
        \includegraphics[width=.25\linewidth]{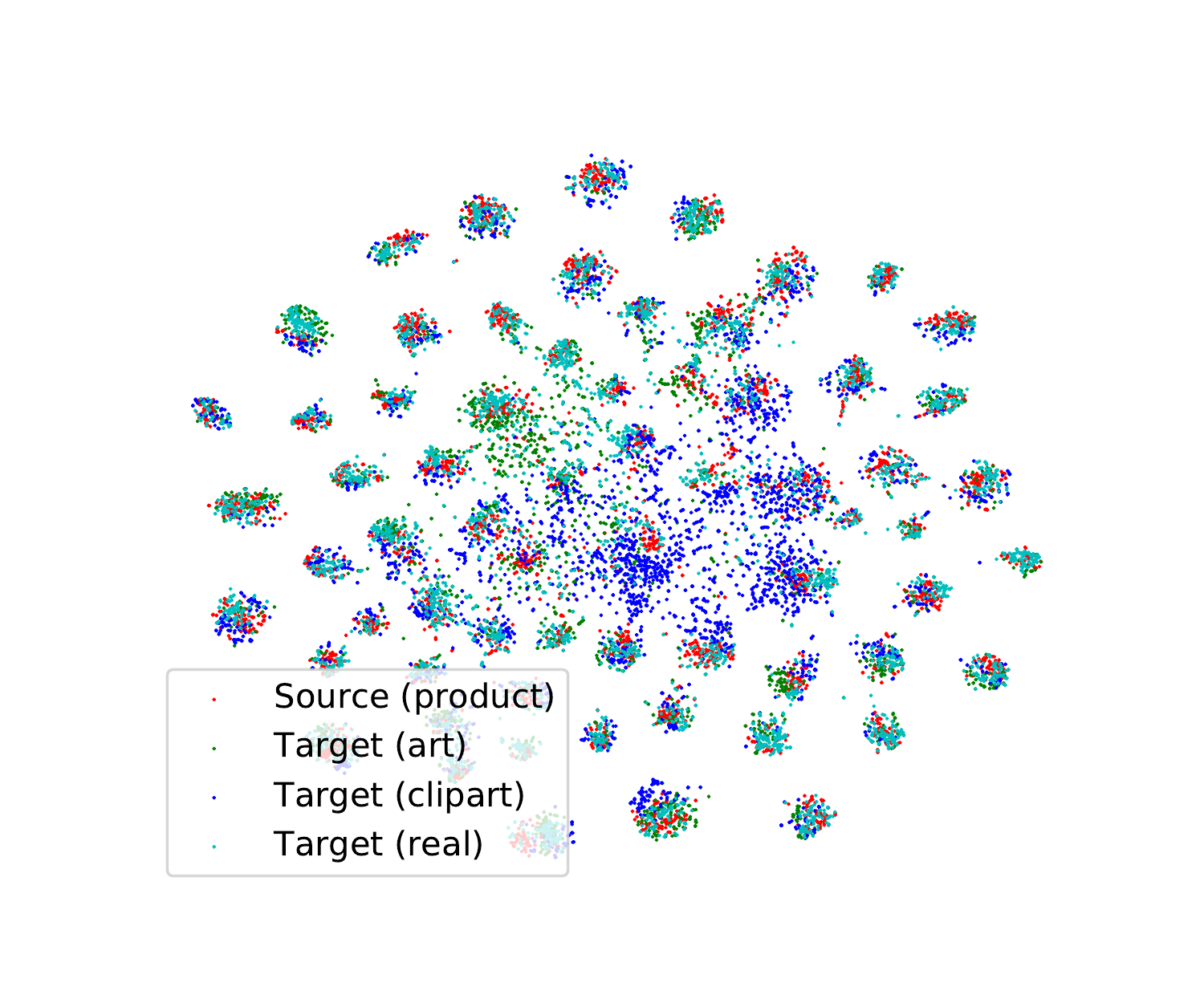} &
        \includegraphics[width=.25\linewidth]{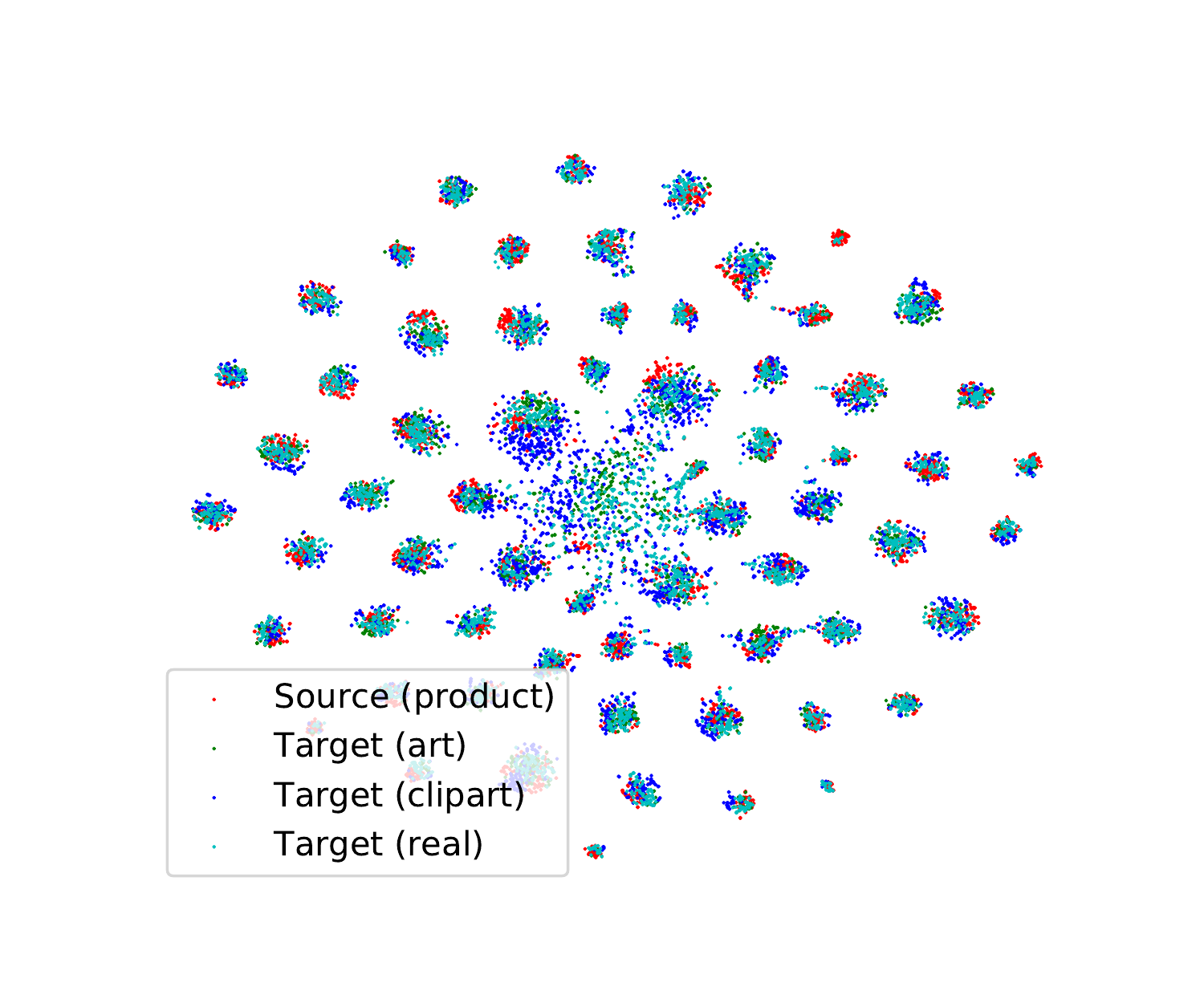} \\
        (a) CDAN~\cite{long2018conditional} & (b) CDAN (\textbf{w}/ domain labels) & (c) CGCT & (d) D-CGCT \\
    \end{tabular}
    \vspace{-.05in}
    \captionof{figure}{\textit{t}-SNE plots of the feature embeddings for the Product $\rightarrow$ \textit{rest} of the Office-Home. All the models use ResNet-50 as  backbone. Each colour indicates a different domain.}
    \label{tab:tsne2}
\end{table*}

\begin{table*}[!t]
    \centering
    \setlength{\tabcolsep}{.2pt}
    \begin{tabular}{c c c c}
        \includegraphics[width=.25\linewidth]{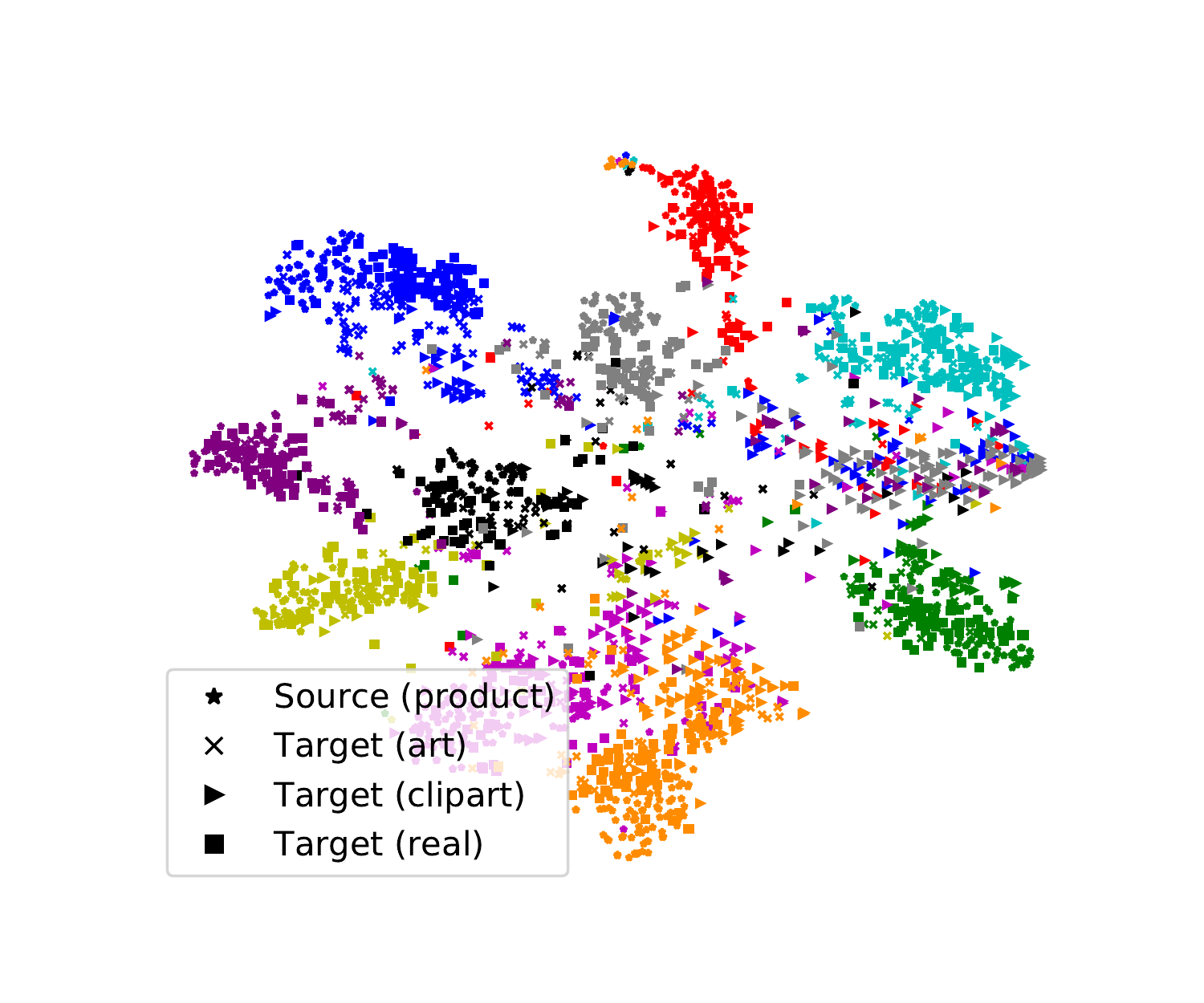} &
        \includegraphics[width=.25\linewidth]{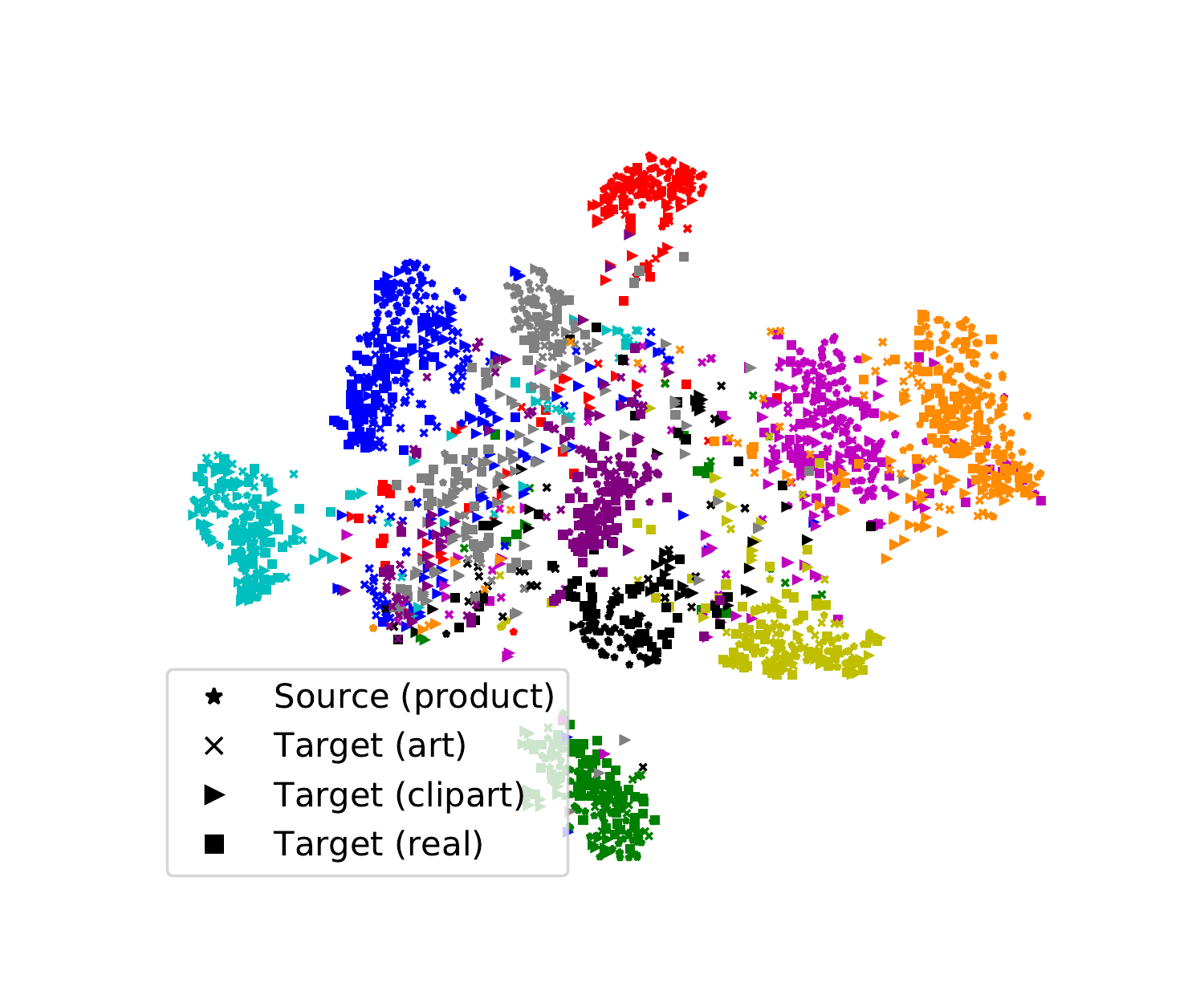} &
        \includegraphics[width=.25\linewidth]{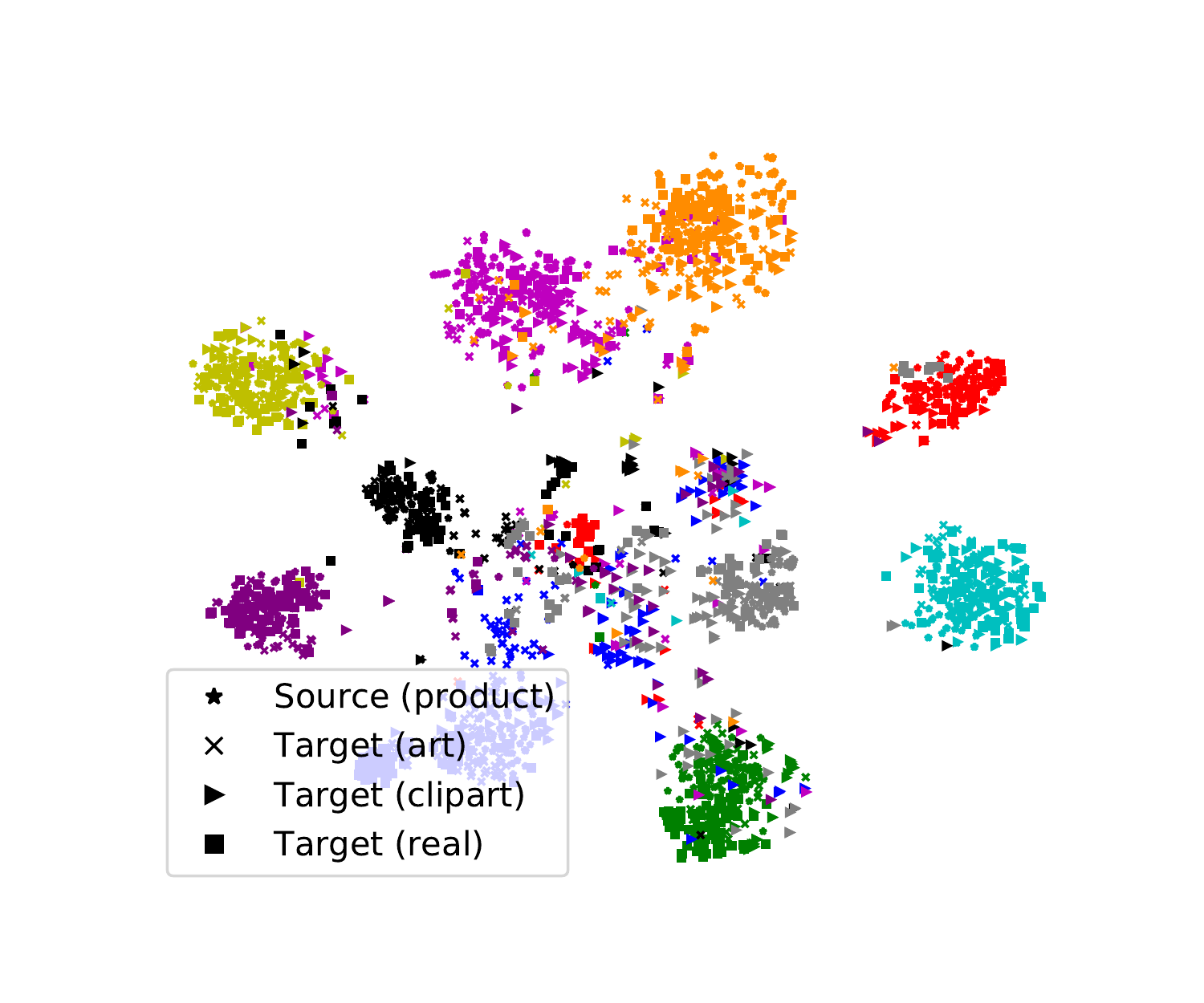} &
        \includegraphics[width=.25\linewidth]{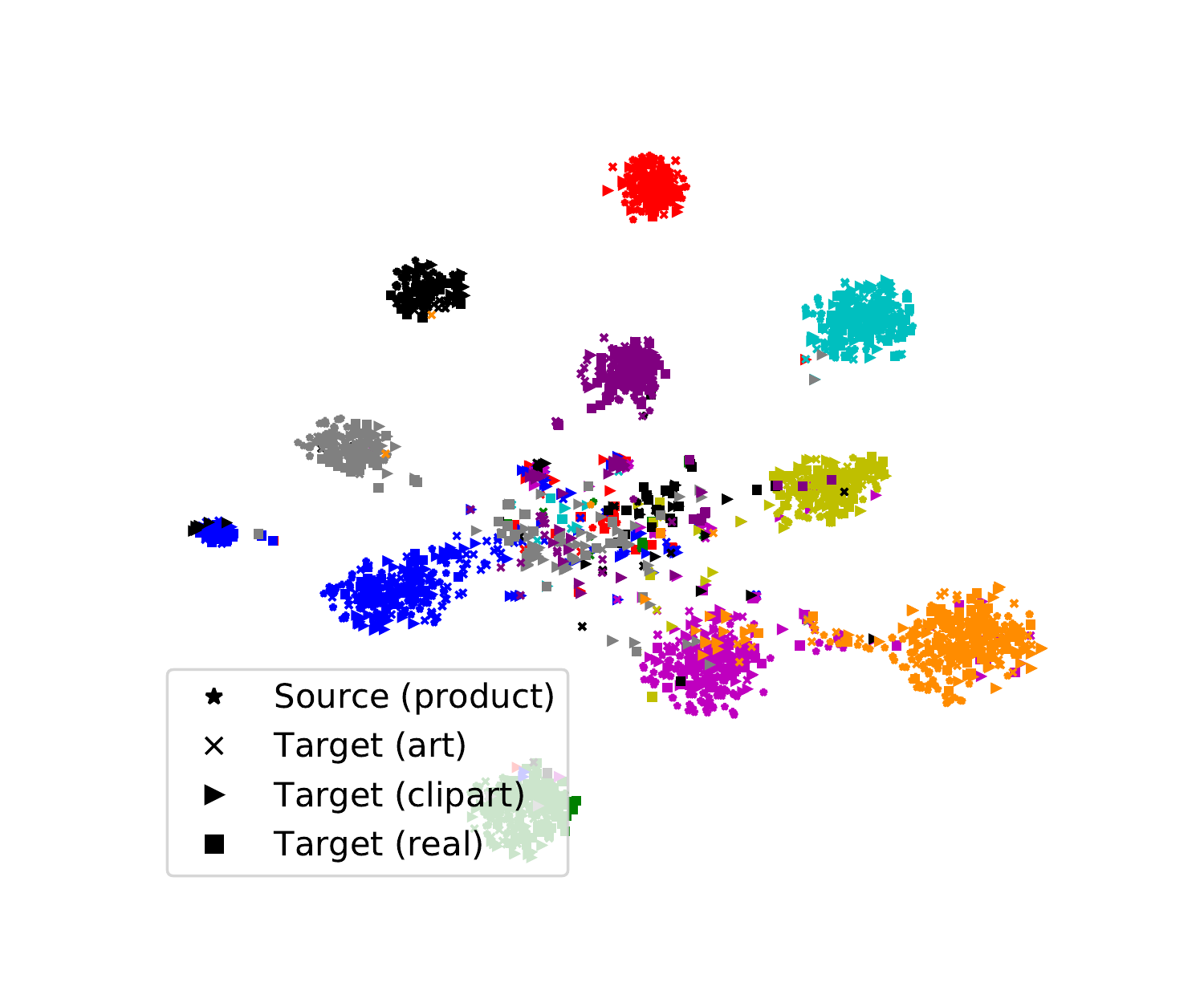} \\
        (a) CDAN~\cite{long2018conditional} & (b) CDAN (\textbf{w}/ domain labels) & (c) CGCT & (d) D-CGCT \\
    \end{tabular}
    \vspace{-.05in}
    \captionof{figure}{\textit{t}-SNE plots of the feature embeddings for the Product $\rightarrow$ \textit{rest} of the Office-Home depicting only 10 randomly sampled classes. All the methods use ResNet-50 as backbone. Each colour indicates a different class while each shape represents a different domain.}
    \label{tab:tsne3}
\end{table*}

In this section we compare with the state-of-the-art methods for the Digits-five and PACS. Since, the recent work of MTDA, HGAN~\cite{yang2020heterogeneous}, does not report results with all the domains available in the PACS and the DomainNet, we additionally report the results with those selected domains in this section for a fair comparison. In the Tab.~\ref{tab:SOTA-digits},~\ref{tab:SOTA-PACS} and~\ref{tab:SOTA-hgan-domainnet}, we club the baselines into two distinct settings: target combined and multi-target. In the former setting, the domain labels of the targets are latent, and all the target domains are combined into a single target domain. Whereas in the latter, each target domain is treated separately. For both the settings, we just train one single model for a given \textit{source} $\rightarrow$ \textit{rest}, as in HGAN~\cite{yang2020heterogeneous}.

In the Tab.~\ref{tab:SOTA-digits}, we report the state-of-the-art comparison on the Digits-five. For a fair comparison, we compare with the baselines reported in~\cite{chen2019blending} that use a backbone network similar to the one described in the Tab.~\ref{tab:digits_arch}. In both the target combined and multi-target settings, our proposed methods outperform all other baselines. For the PACS, reported in the Tab.~\ref{tab:SOTA-PACS}, we notice that domain labels is very vital for mitigating multiple domain-shifts. For example, CDAN in the multi-target setting performs 9.8\% better than its target combined counterpart. Similar trend can also be observed between our \ours and \domours, with the \domours outperforming the former by a large margin. Finally, we re-evaluate our methods on the 5 domains of the DomainNet, by leaving out the Quickdraw domain as in~\cite{yang2020heterogeneous}. Results are reported in the Tab.~\ref{tab:SOTA-hgan-domainnet}. We produce state-of-the-art performance in the DomainNet for both the settings by non-trivial margins. This further shows that our proposed feature aggregation and training strategy are much more effective than the HGAN.


\subsection{Visualization}
In this section we visualize the features learned by our models and compare them with the baseline methods. The Fig.~\ref{tab:tsne2} depicts the \textit{t}-SNE plots of the feature embeddings computed by feature extractor network (ResNet-50) for the direction Product $\rightarrow$ \textit{rest} of the Office-Home. The plots in the Fig.~\ref{tab:tsne2} (c) and (d) demonstrate that the proposed \ours and \domours result in well clustered and discriminative features compared to CDAN baselines (see Fig.~\ref{tab:tsne2} (a) and (b)). To better visualize the decision boundaries in the latent feature space, we select 10 classes, randomly from the Office-Home, and depict the \textit{t}-SNE plots of the feature embeddings in the Fig.~\ref{tab:tsne3}. It is can be seen that our models learn features that can be easily separated by a linear classifier, much easier than the CDAN models. In particular, the CDAN when using domain labels (see Fig.~\ref{tab:tsne3} (b)) produces more overlapping classes than our \domours (see Fig.~\ref{tab:tsne3} (d)). Thus, when the domain labels are leveraged with our DCL strategy, the model produces features that are more discriminative, thereby leading to an improved performance in the MTDA.
